# Feasibility of a Patient-Specific Digital Twin for Ventriculoperitoneal Shunt Trajectory Planning

Kritsanavis Chongsrid[1,2,3,4,*], Vich Yindeedej[5,*], Supaporn Konmun[5], Xulei Yang[4,†] & Si Yong Yeo[1,2,3,†]

[1]MedVisAI Lab, Singapore. [2]Lee Kong Chian School of Medicine, Nanyang Technological University, Singapore. [3]Centre of AI in Medicine, Singapore. [4]A*STAR Institute of Advanced Intelligence and Computing (A*STAR IAIC), Agency for Science, Technology and Research (A*STAR), Singapore. [5]Division of Neurosurgery, Department of Surgery, Faculty of Medicine, Thammasat University Hospital, Thammasat University, Pathum Thani, Thailand.

[*]These authors contributed equally: Kritsanavis Chongsrid, Vich Yindeedej.
[†]Corresponding authors. E-mail: yang_xulei@a-star.edu.sg; siyong.yeo@ntu.edu.sg

## Abstract

Catheter obstruction is the commonest cause of shunt failure; freehand insertion at Keen's point leaves the tip free in cerebrospinal fluid in only one case in ten. We built a digital twin that reconstructs skull, scalp, ear and ventricles from preoperative CT by deterministic geometry, searches every entry around Keen's point for the perpendicular trajectory with the longest intraventricular length under safety constraints, and returns the plan as two ruler measurements from the ear and two depths. Executed freehand in 11 consecutive adults (IDEAL Stage 2a), it achieved Hayhurst Grade 1 placement in 10 of 11 against 22 of 213 historically ($p = 9 \times 10^{-9}$), median entry and angular deviations 2.5 mm and 5.9°, 1.4 cm more intraventricular catheter than Keen's point, and no intraoperative complication. The technique was unchanged; the gain came from choosing entry and depth per patient. Whether it lowers revision is the next stage's hypothesis.

## Introduction

Ventriculoperitoneal (VP) shunt placement is the most common operation for hydrocephalus[1,2], and it fails often. About a third of adults require revision, most within the first six months[3], and the commonest cause is obstruction of the ventricular catheter[4–7]. The catheter's position determines that risk. Tips lying in the anterior part of the ipsilateral ventricle, surrounded by cerebrospinal fluid and clear of the septum and ventricular wall, fail less often[5,8,9], because the perforated inlet segment extends 16 to 24 mm from the tip and occludes wherever it rests against ependyma, choroid plexus or brain[7,10,11]. The whole segment must therefore lie free in cerebrospinal fluid, a length that undilated ventricles rarely offer and that a dilated ventricle offers only along its body, the one straight segment, entered from the atrium[10]. Where the catheter enters and where it lies are settled in the operating room by an entry point read off skin landmarks and a direction held by hand[1].

In standard practice the surgeon chooses the route based on the patient's anatomy and pathology: frontal, posterior parietal or occipital. Within the route, however, the entry point is a fixed offset from an external landmark, Kocher's point, Keen's point or Frazier's point[12,13], not adapted to the individual ventricle or skull, although the entry point largely determines where the catheter ends and how much parenchyma it crosses[14,15]. The catheter is then passed perpendicular to the skull[16,17]. Refinements of the offsets[14,18–20], per-patient choice of the entry on the CT[21,22] and an in silico map of the posterior skull from which a perpendicular reaches the ventricle[23] have each improved on the landmark without giving the surgeon a point, a direction and a depth chosen for the patient on an explicit criterion. Freehand ventriculostomy achieves optimal placement in about half of cases[24–26]. The posterior parietal route, used for most shunts at the study institution and in routine use across Asia[27,28], performed supine and needing no incision beyond those the shunt requires, is where freehand accuracy is lowest[1,27,29] and where revision is no more frequent than after frontal entry[1,30,31]. Angle guides hold a direction measured on the preoperative image but leave the burr hole to the landmark[32]. Neuronavigation and ultrasound improve placement in individual series[33,34], but meta-analyses show little or no effect on revision[35,36]. Navigation, ultrasound, augmented reality and robotic assistance also address execution rather than planning. They show where the catheter lies or hold it on a trajectory the surgeon has specified[37–39], and no step evaluates candidate trajectories against the patient's own ventricle. They also require equipment, time and training that many centres lack, so that freehand insertion remains the predominant method[1].

A digital twin is a patient-specific replica that predicts outcomes and informs decisions[40,41] and, in surgery, lets surgeons rehearse an access route before taking it[42]. A static twin built by segmenting the preoperative image is the form in established clinical use[43,44]. It has been applied in neurosurgery to resection, deep brain stimulation and vascular intervention[45,46]; to trajectory planning for haematoma evacuation[47]; and, in endovascular surgery, to rehearsing the deployment of a device in the patient's preoperative CT[48,49]. We built one for VP shunt planning. From the routine preoperative CT it searches the posterior parietal skull around Keen's point for the perpendicular entry that maximises the catheter's continuous length within the ipsilateral ventricle under explicit safety constraints. It returns the plan as the two measurements from the pinna that define Keen's point, together with the depths to the ventricle and within it, so that the surgical team can execute it with a ruler. The plan is a recommendation that the surgical team reviews before surgery and decides whether to follow. The planner is deterministic geometry rather than a learned model, an engineering decision for deployment. It needs no graphics processor, training data or network connection, returns the same plan for the same scan, and leaves every intermediate result open to inspection. The plan needs no intraoperative hardware, which matters most where the burden of hydrocephalus is greatest and navigation rarest[50]. These properties are among the priorities that an expert consensus on artificial intelligence in surgery sets for its adoption: transparency, integration into the existing workflow and scalability across centres with different resources[51]. A systematic review of digital twins in neurosurgery found 25 comparative studies to 2022, none in shunt surgery[52], and to our knowledge none has been evaluated in patients.

We report the plan's technical feasibility, safety and execution fidelity, at IDEAL Stage 2a (Development)[53], in 11 consecutive adults at one centre in whom it was executed freehand for posterior parietal VP shunt insertion. The primary outcome is the proportion of Grade 1 catheter placements on immediate postoperative CT, against the institution's freehand Keen's point series as historical control; secondary outcomes are execution fidelity, proximity to critical structures and a geometric comparison with the classical and the modified Keen's point, each constructed by its definition on the same scans. It is the development stage before multicentre validation and a comparative trial of whether trajectory planning reduces shunt failure.

## Results

### Patient flow and characteristics

Eleven consecutive patients had the plan executed and were analysed, 7 male and 4 female, median age 62 years (range 43–82). Of the 17 patients scheduled for the operation in the study period, 12 were planned and 11 had the plan executed. Two of the other five were outside the conditions for the plan (Methods), one with an Evans index of 0.57, above the limit of 0.45, and one with a pinna folded against the head on the preoperative CT. In three, the surgical team chose another course on clinical grounds, and in one planned patient the side of insertion was changed at surgery because of a skin infection over the planned entry. Indications for VP shunt placement included normal pressure hydrocephalus in 2 patients (18%), intracranial hemorrhage (cerebellar, pontine, and thalamic) in 3 patients (27%), ruptured aneurysm (ICA and anterior communicating artery) in 2 patients (18%), brain tumor (GBM, meningioma) in 2 patients (18%), cerebellar AVM in 1 patient (9%), and brain metastasis in 1 patient (9%). The catheter was placed on the right in 10 patients and on the left in one (Case 3), in whom a recurrent right temporo-parietal glioblastoma occupied the posterior parietal corridor and had distorted the right hemisphere and ventricle, so the surgical team elected the left side before planning (Methods). Six patients (55%) had prior shunt placement: 3 had left frontal shunts, 1 had a right frontal shunt, 1 had bilateral frontal shunts, and 1 had bilateral frontal burr holes without visible shunt track. The median Evans index was 0.32 (range 0.28 to 0.40), and the median maximal frontal horn width was 43.6 mm (range 38.6 to 55.2).

## Plan generation

The system returned a plan for each of the 11 patients (Figure 1). Computation time was approximately 15 minutes per case on the CPU and 5 minutes with the GPU (Supplementary Note 2). All generated trajectories met predefined safety constraints on independent re-measurement against the preoperative ventricle: parenchymal traversal $D_1$ ≤4.0 cm (range 2.1–3.7 cm), intraventricular catheter length $D_2$ ≥2.5 cm (range 3.2–5.3 cm), insertion depth D ≥6.0 cm (range 6.4–7.6 cm).

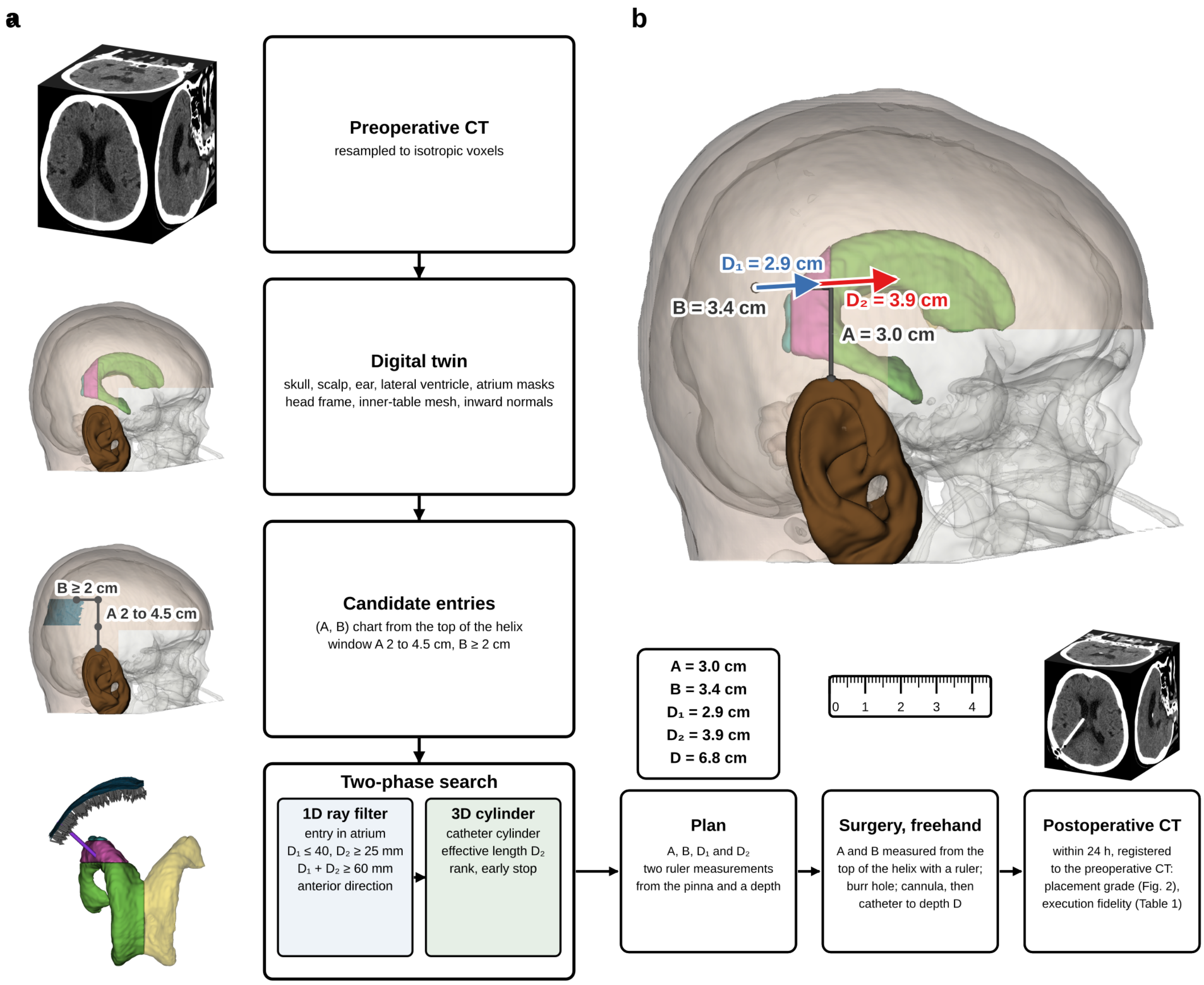

Figure 1. The planning system and the plan as delivered to the surgical team. (a) Pipeline from the preoperative CT to the plan: segmentation, alignment to the head frame, the inner-table mesh, candidate entries parameterised by (A, B) from the top of the helix, the one-dimensional ray filter and the three-dimensional cylinder check, the output A, B, $D_1$ and $D_2$, and the surgical and imaging steps that follow. Thumbnails and panel (b) show Case 6, with the scalp removed where it lies anterior to and below the skull centroid: ipsilateral ventricle green with the atrium pink and the occipital horn teal, contralateral ventricle yellow, pinna brown, candidate window blue, candidate normals grey, planned trajectory

purple. Above the workflow boxes: the plan as delivered, the ruler, and the postoperative CT of the same patient at the same level, with the catheter. (b) The plan on the rendered head of Case 6: A vertically from the top of the helix, B posteriorly along the scalp to the entry, and the perpendicular trajectory with $D_1$ (light blue) through parenchyma and $D_2$ (red) within the ventricle, D = 6.8 cm; ipsilateral ventricle green with the atrium pink and the occipital horn teal, pinna brown.

### Primary outcome: catheter placement accuracy

On the Hayhurst classification[54], Grade 1 placement was achieved in 10 of 11 cases (90.9%, 95% CI 62.3–98.4%; Figure 2). Grade 2 placement occurred in 1 of 11 cases (9.1%), a 60-year-old man with a cerebellar arteriovenous malformation (Case 2) in whom the catheter tip displaced the septum pellucidum without crossing it. No Grade 3 placements occurred. By tip location, the tip lay in the ipsilateral body or frontal horn of the lateral ventricle in 10 of 11 cases (90.9%, 95% CI 62.3–98.4%) and at the septum pellucidum in one (Case 2, the Grade 2 placement); no tip lay in the third ventricle, contralateral ventricle or parenchyma.

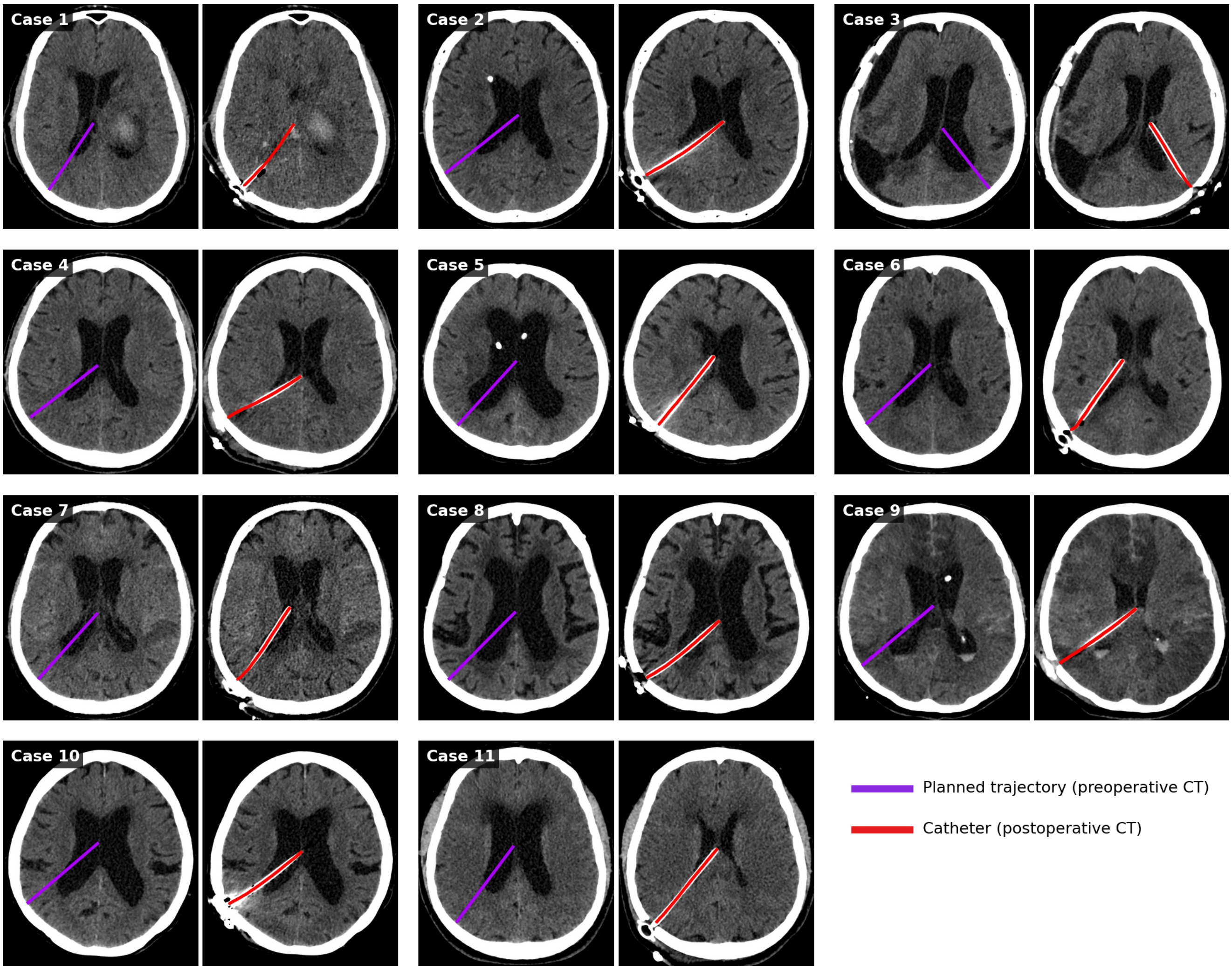

Figure 2. Planned and executed trajectories in all 11 cases. For each case, the preoperative CT with the planned trajectory (purple, left) and the postoperative CT, registered to it, with the catheter (red, right), both shown in the same reformatted plane, so that the skull is identical on the two sides apart from the burr hole. Placement was Hayhurst Grade 1 in all cases except Case 2 (Grade 2). Deviations are given in Table 1 and the two trajectories are shown in three dimensions in Figure 3.

### Execution fidelity

### Table 1. Deviation Between Planned and Executed Trajectories

| Metric | Median | IQR | Range |
|---|---|---|---|
| Entry point deviation, mm | 2.5 | 1.3–4.9 | 0.4–6.9 |

| Metric | Median | IQR | Range |
|---|---|---|---|
| Tip position deviation, mm | 7.0 | 4.6–7.3 | 3.7–8.4 |
| Catheter length deviation, mm | 1.3 | 0.5–2.6 | 0.1–3.9 |
| Angular deviation | | | |
| Geometric, ° | 10.5 | 8.9–14.5 | 6.5–20.8 |
| Execution, ° | 9.5 | 5.9–15.8 | 3.5–21.0 |
| Composite, ° | 5.9 | 4.4–7.6 | 3.6–12.2 |

Table 1 gives the deviations between planned and executed trajectories and Figure 3 shows the two trajectories of every case in three dimensions. Supplementary Table 1 gives the direction of each trajectory (azimuth and elevation) and Supplementary Figure 4 illustrates these two angles.

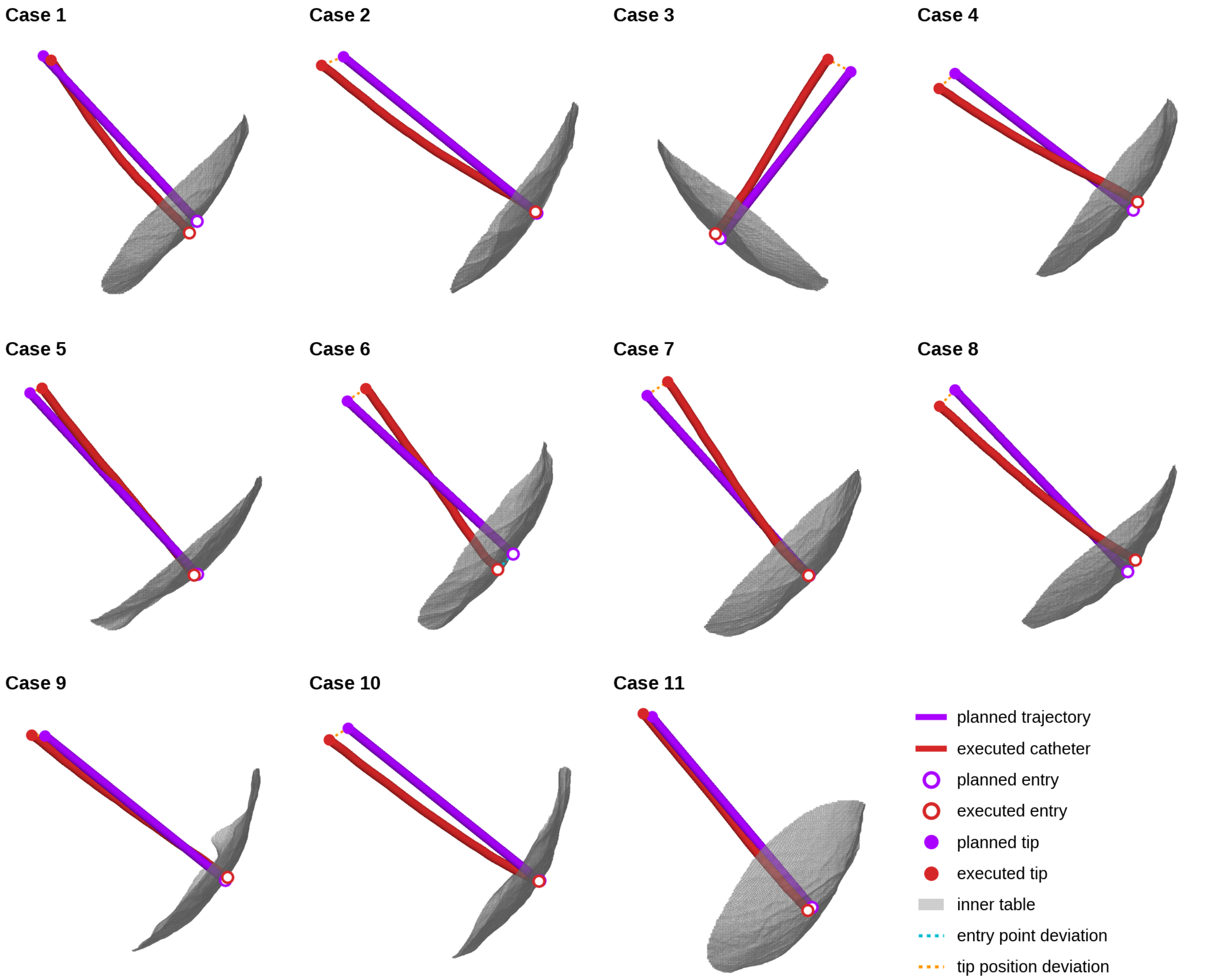

Figure 3. Planned and executed trajectories in three dimensions, all 11 patients. Each panel is viewed from above at one common scale, anterior up, the patient's right on the right. The inner table within 35 mm of the planned entry (grey) is from the preoperative CT. The planned trajectory is purple; the executed catheter, traced on the postoperative CT registered to the preoperative CT, is red. Open circles mark the planned entry and the proximal end of the catheter trace, dots the tips; dotted lines join the two entries (cyan) and the two tips (orange), the entry point and tip position deviations of Table 1.

Nine of 11 cases (82%) achieved entry point deviation <5 mm. Six of 11 cases (55%) achieved execution angular deviation <10°. The composite deviation (median 5.9°, IQR 4.4–7.6°) was smaller than either of its components, the geometric (10.5°) and the execution (9.5°), in the cohort medians and in 7 of 11 individual cases.

### Safety outcomes

No intraoperative complications occurred. The executed catheter measured a median 7.2 cm (IQR 6.8–7.6) against a planned insertion depth of 7.3 cm (IQR 6.8–7.5), none shorter than 6.0 cm.

Figure 4 gives the planned and executed minimum distances to 47 structures; four further structures lay on the intended catheter path and had a median distance of 0 mm for both planned and executed trajectories: CSF, ipsilateral cerebral white matter, ipsilateral lateral ventricle, and ipsilateral inferior parietal lobule. No executed catheter intersected any other structure. No planned or executed trajectory crossed the temporalis muscle; the median distance from the extracranial course of the trajectory to the muscle was 21 mm (range 10–52) planned and 20 mm (range 11–50) executed.

Of the ventricular wall structures, the ipsilateral thalamus was the closest to the catheter (median 3.6 mm planned, 2.2 mm executed), followed by the ipsilateral caudate (7.2 mm, 4.1 mm); the ipsilateral hippocampus (16.3 mm, 16.2 mm), contralateral thalamus (7.6 mm, 8.2 mm) and contralateral caudate (18.5 mm, 15.8 mm) were further. No executed catheter contacted any of them. Across the remaining structures the median executed distance differed from the planned distance by less than 4 mm in every row (Figure 4).

All catheters entered CSF in the ipsilateral lateral ventricle. Postoperative imaging showed no haemorrhage along the track, and no catheter tip contacted the choroid plexus or penetrated the ependyma; the only contact with the ventricular wall was the Case 2 tip against the septum pellucidum, the Grade 2 placement reported above.

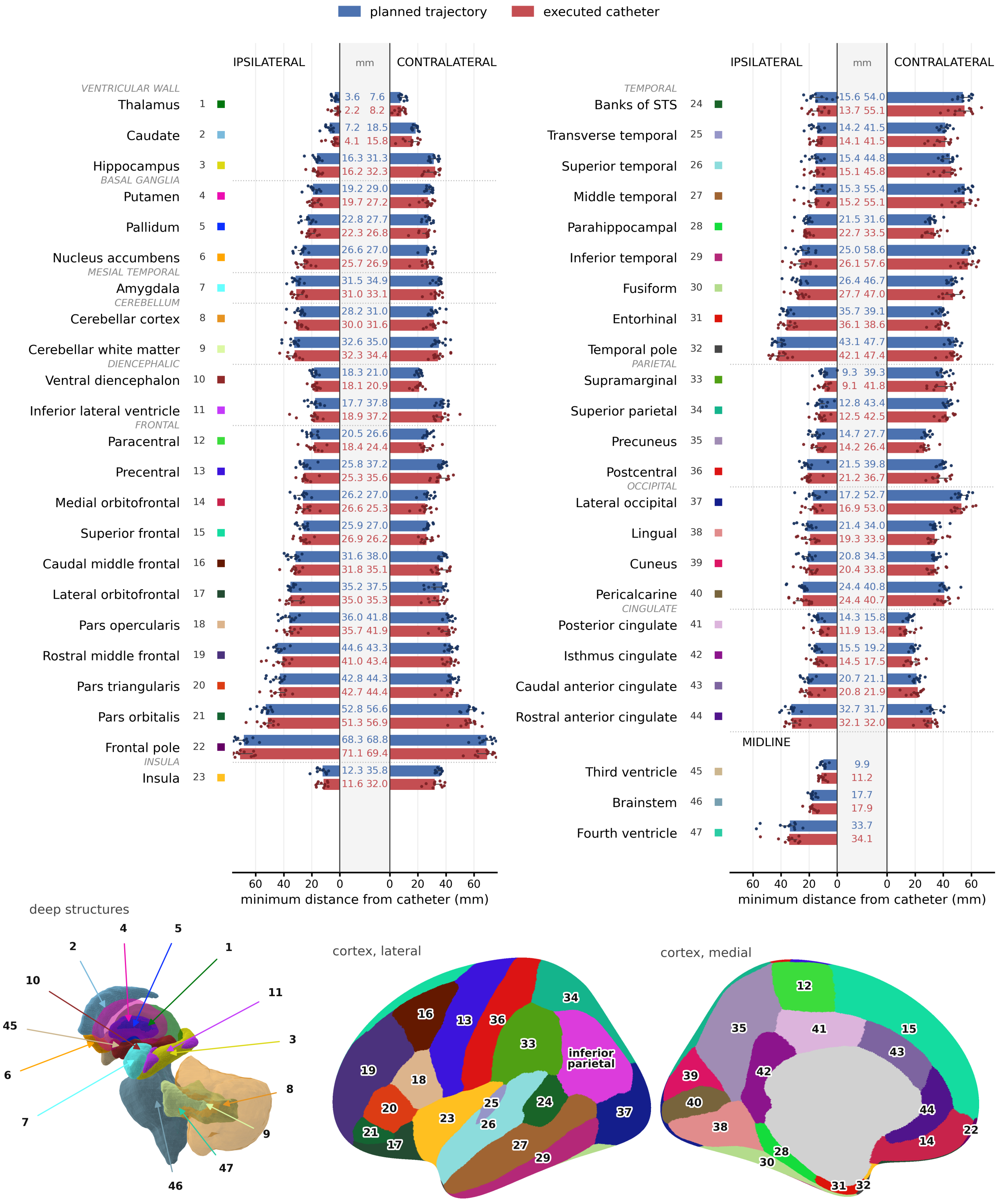

Figure 4. Minimum distance from the catheter to 47 structures, planned versus executed. Bars are medians with interquartile whiskers and dots the individual patients; blue, planned trajectory; red, executed catheter. Numbers identify each structure on the atlas below; the three midline structures are plotted separately, and the inferior parietal parcel through which the catheter passes is labelled on the atlas and has no row.

## Geometric counterfactual analysis

Table 2 compares the plan with the classical Keen's point trajectory on the same CT, and Figure 5 draws the plan and both landmark trajectories for every patient.

**Table 2. Algorithm-Guided vs. Keen's Point Trajectory Constructed by Its Definition, at 6 cm and 8 cm: Per-Case Comparison**

| Case | Algorithm | | | Keen's point, 6 cm | | | Keen's point, 8 cm | | |
|---|---|---|---|---|---|---|---|---|---|
| | $D_1$ | $D_2$ | Overshoot | $D_1$ | $D_2$ | Overshoot | $D_1$ | $D_2$ | Overshoot |
| 1 | 3.4 | 3.4 | 0.0 | 3.6 | 2.3 | 0.1 | 3.6 | 2.3 | 2.1 |
| 2 | 3.3 | 4.2 | 0.0 | 3.2 | 2.8 | 0.0 | 3.2 | 3.5 | 1.3 |
| 3 | 2.4 | 4.0 | 0.0 | 2.7 | 3.3 | 0.0 | 2.7 | 3.6 | 1.7 |
| 4 | 3.6 | 3.2 | 0.0 | 3.7 | 0.8 | 1.5 | 3.7 | 0.8 | 3.5 |
| 5 | 2.1 | 5.3 | 0.0 | 2.3 | 3.7 | 0.0 | 2.3 | 5.5 | 0.3 |
| 6 | 2.9 | 3.9 | 0.0 | 2.5 | 2.5 | 1.0 | 2.5 | 2.5 | 3.0 |
| 7 | 2.9 | 4.4 | 0.0 | 3.0 | 2.7 | 0.3 | 3.0 | 2.7 | 2.3 |
| 8 | 2.7 | 4.8 | 0.0 | 3.4 | 2.6 | 0.0 | 3.4 | 3.4 | 1.2 |
| 9 | 2.8 | 4.1 | 0.0 | 3.6 | 2.4 | 0.0 | 3.6 | 3.0 | 1.4 |
| 10 | 3.1 | 4.3 | 0.0 | 2.9 | 3.1 | 0.0 | 2.9 | 3.1 | 2.0 |
| 11 | 3.7 | 3.9 | 0.0 | 3.2 | 2.8 | 0.0 | 3.2 | 3.4 | 1.4 |
| Median | 2.9 | 4.1 | 0.0 | 3.2 | 2.7 | 0.0 | 3.2 | 3.1 | 1.7 |
| IQR | 2.8–3.4 | 3.9–4.4 | 0.0–0.0 | 2.8–3.5 | 2.5–2.9 | 0.0–0.2 | 2.8–3.5 | 2.6–3.5 | 1.3–2.2 |

All values are in centimetres. Parenchymal ($D_1$) = inner skull table to first ventricle contact; intraventricular ($D_2$) = length of catheter from ventricular entry to first exit from the ipsilateral lateral ventricle; overshoot = catheter beyond that length, including any length in the contralateral ventricle after crossing the septum pellucidum. The landmark arm assumes the entry and angle are achieved exactly. The Keen's point line is the same at both depths, so parenchymal traversal is identical. Constraints: $D_1$ ≤4.0 cm, $D_2$ ≥2.5 cm, D ≥6.0 cm.

Algorithm-generated trajectories achieved greater intraventricular length than the Keen's point trajectory at 6 cm in every case (median difference 1.4 cm, 11/11; Wilcoxon signed-rank p=0.0010). The difference persisted when the Keen's point trajectory was inserted to the same depth as the algorithm plan rather than a fixed 6 cm (median advantage 0.8 cm, 11/11). No algorithm trajectory extended beyond the ventricle, whereas 4 of 11 Keen's point trajectories at 6 cm did, reaching 1.5 cm (Cases 1, 4, 6, 7). Four Keen's point trajectories also left and re-entered the ventricle, placing the mid-portion of the catheter against ependyma or parenchyma; no algorithm trajectory did so. At 8 cm, the Keen's point trajectory achieved a median intraventricular length of 3.1 cm (IQR 2.6–3.5; median difference from the algorithm 1.1 cm, 10/11; Wilcoxon p=0.0020), extended beyond the ipsilateral ventricle in 11 of 11 cases (median overshoot 1.7 cm, maximum 3.5 cm), and crossed the septum pellucidum into the contralateral ventricle in 9 of 11. Parenchymal traversal did not differ between the two approaches (median paired difference −0.2 cm, Wilcoxon p=0.46).

The Keen's point trajectory at 6 cm fell below the 2.5 cm intraventricular floor in 3 of 11 cases (Cases 1, 4 and 9, at 2.3, 0.8 and 2.4 cm), and at 8 cm in 2 of 11 (Cases 1 and 4).

For the modified Keen's point of Colombo et al.[19] (Supplementary Table 3), the entry lay 1.8 to 3.4 cm below the classical Keen's point (median 2.5 cm), over the posterior temporal bone. From that entry the perpendicular runs medially rather than along the body, so the intraventricular length was a median 1.4 cm (IQR 1.2–2.0), against 2.7 cm for classical Keen's and 4.1 cm for the algorithm (algorithm versus modified Keen's, median difference 2.3 cm, 11/11, Wilcoxon signed-rank p=0.0010; classical versus modified Keen's, 1.1 cm, 9/11, p=0.0049). The tip lay outside the ipsilateral ventricle in 10 of 11 cases at both 5 and 6 cm.

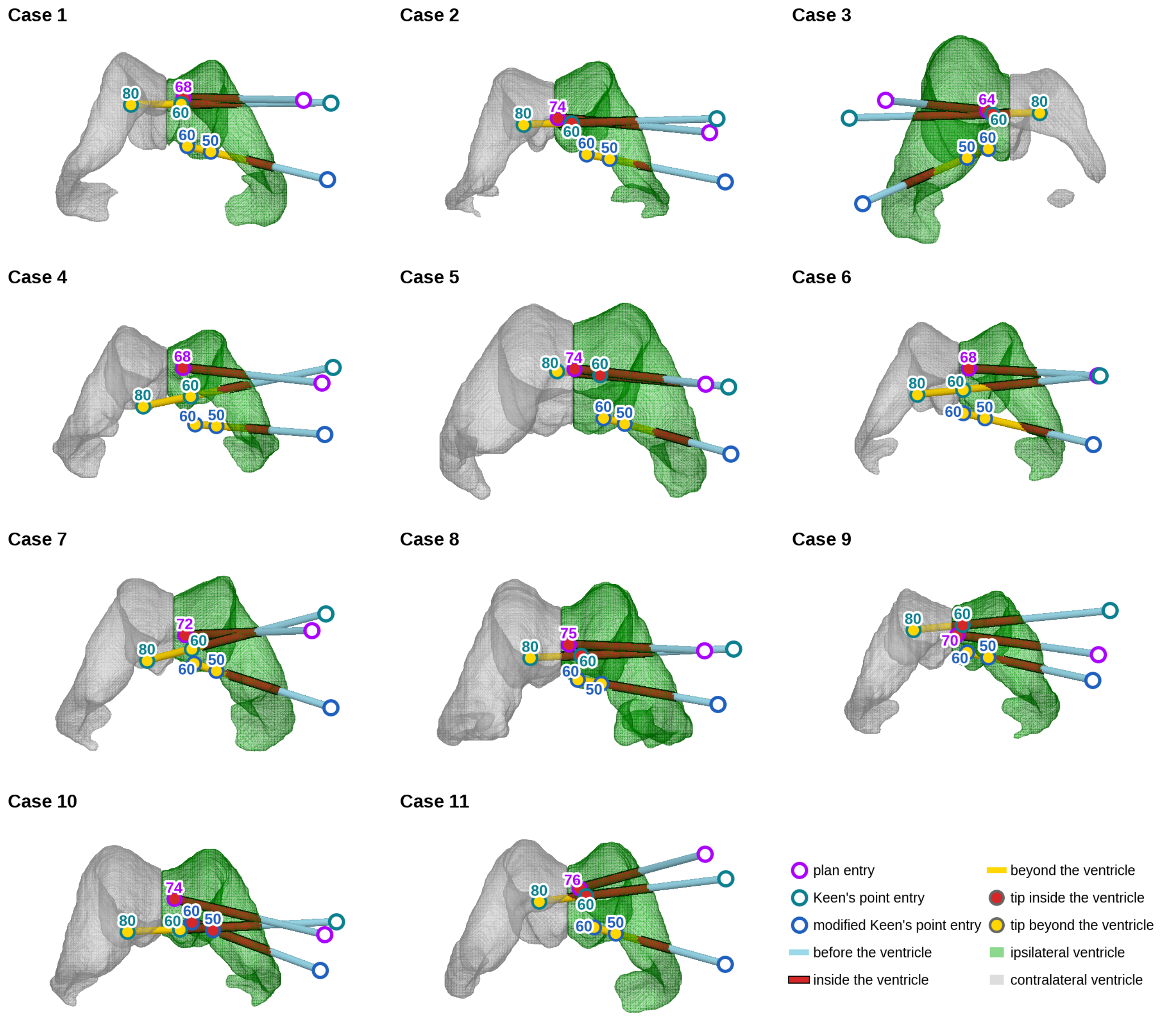

Figure 5. The computed plan and the two landmark trajectories in all 11 patients. Each panel is viewed from behind at one common scale, the patient's right on the right. The ipsilateral lateral ventricle is green and the contralateral grey (SynthSeg segmentation). Trajectories: the computed plan (purple); the classical Keen's point at 6 and 8 cm (teal); the modified Keen's point[19] at 5 and 6 cm (blue). Each trajectory is light blue before the ventricle, red inside it and outlined in black, and yellow beyond it; tips are dots labelled with their depth in millimetres, the edge of the dot and its label in the colour of the entry. Lengths and overshoots are given in Table 2 and Supplementary Table 3.

## Comparison to historical controls

Figure 6 compares the placement grade distribution of the present cohort with the freehand Keen's point cohort of the same institution[1], in which Grade 1 placement was achieved in 22 of 213 cases (10.3%), Grade 2 in 182 (85.5%) and Grade 3 in 9 (4.2%). Grade 1 placement was 90.9% (10 of 11) here against 10.3% historically ($p = 9 \times 10^{-9}$, Fisher's exact test). On the tip-location classification used in that series, a good location (ipsilateral frontal horn or body) was achieved in 10 of 11 algorithm-guided cases versus 41 of 213 (19.2%) historical Keen's point cases ($p = 1.6 \times 10^{-6}$, Fisher's exact test), the remaining tip lying at the septum pellucidum against 100 of 213 (46.9%) historically (Figure 6b). Entry accuracy was 0.3 ± 0.2 cm (range 0.0–0.7) against 1.6 ± 1.0 cm (0–5) in that series (Figure 6c), with catheter lengths of 7.2 ± 0.5 cm (6.3–7.8) against 8.0 ± 1.2 cm (5–11) (Figure 6d).

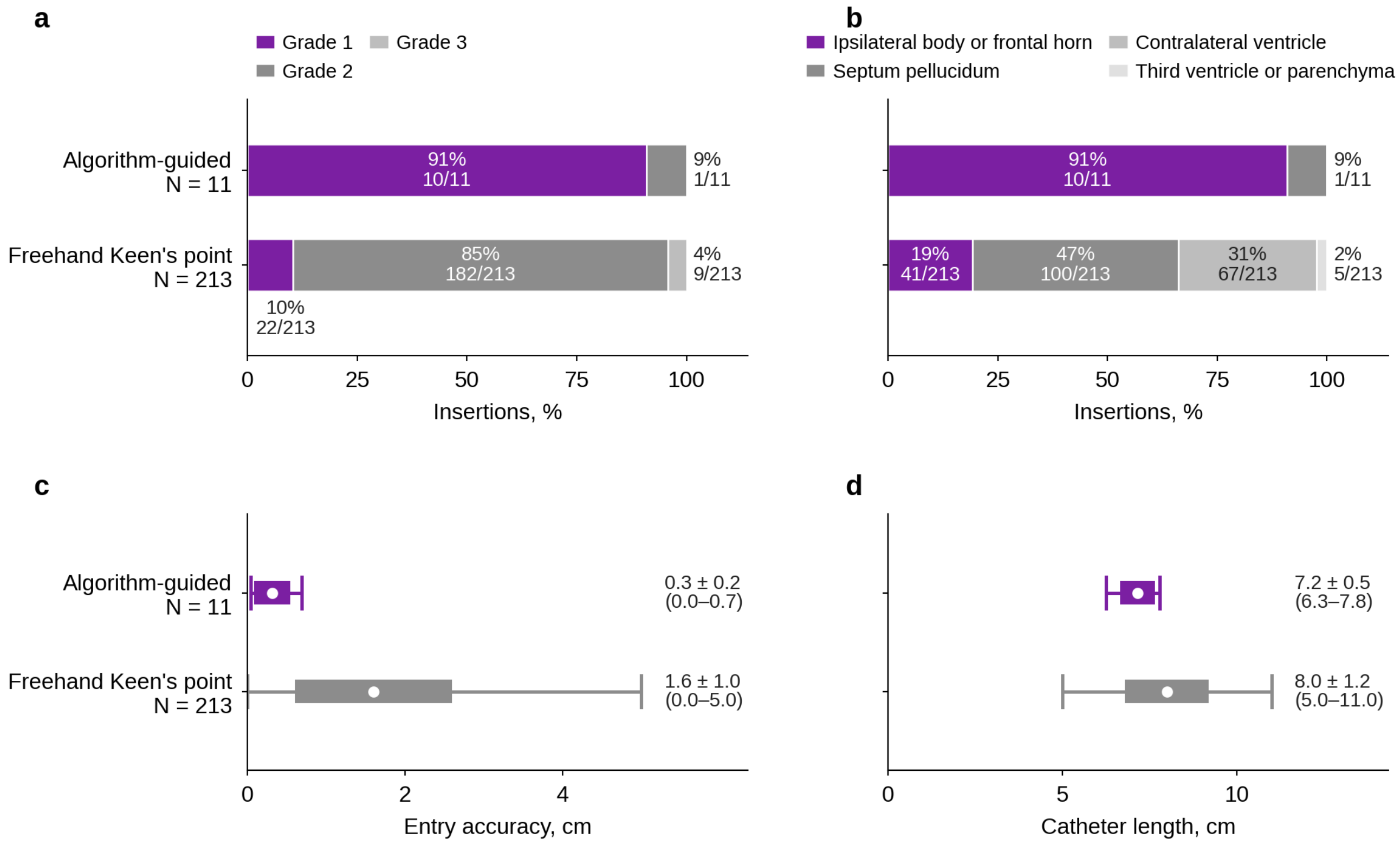

Figure 6. The present cohort against the freehand series of the same institution[1]. (a) Hayhurst grade for the 11 algorithm-guided insertions and the 213 freehand Keen's point insertions; each segment is labelled with its percentage and count. (b) Tip location on the scheme of Chotsakulthong et al.[1] for the same two cohorts; contralateral ventricle pools the contralateral lateral wall, frontal horn and body of that scheme. (c) Entry accuracy, the distance between the intended and the executed entry point, as mean ± SD with the range, for the freehand Keen's point series[1] and the present cohort, in which the intended entry is the planned one. (d) Catheter length, as mean ± SD with the range, for the same two cohorts.

## Discussion

Executed freehand from two ruler measurements and two depths, a ventricular catheter trajectory computed from the preoperative CT alone gave Grade 1 placement in 10 of 11 consecutive patients, against 22 of 213 (10.3%) in the freehand Keen's point series from the same institution[1]. Ten of 11 tips lay in the ipsilateral body or frontal horn, against 19.2% historically, and the plans were executed with a median entry deviation of 2.5 mm and a median angular deviation of 5.9°. The gain over the historical series came from where and how deep the catheter was placed, not from how it was inserted. Both teams inserted perpendicular catheters through a burr hole placed by measurement from the pinna; what changed were the numbers, A and B and the depths $D_1$ and $D_2$, while the measuring, the burr hole and the cannula pass remained in the surgical team's hands. The paired counterfactual in Table 2 isolates the difference. On the same 11 ventricles the computed trajectory achieved a median 1.4 cm more intraventricular catheter than the Keen's point trajectory and met the 2.5 cm floor in every case, whereas the Keen's trajectory failed to meet it in three. It never left the ventricle, which the Keen's trajectory did in four, and parenchymal traversal did not differ. The 8 cm construction shows the other half of the mechanism. At the depth surgeons at this institution actually used, the Keen's point catheter left the ipsilateral ventricle in 11 of 11 and crossed the septum in 9 of 11, the signature of the historical series, in which longer catheters predicted malposition at Keen's point and not at Kocher's[1]. The plan's depths, $D_1$ to the ventricle and $D_2$ within it, are the numbers a landmark cannot supply. The execution data show that the surgical team oriented to the plan rather than to the local skull. The composite angular deviation (5.9°) was smaller than either of its components (9.5° and 10.5°), so a 2.5 mm entry shift on a curved skull was compensated by hand. The execution deviation of 9.5° in patients is close to the 8.6° recorded for the same quantity when surgeons aimed at this route on printed skulls[55], where the head was clamped and the probe tracked. This is the fidelity a plan delivered by ruler can expect from an experienced team, and the quantity a multicentre study should measure first.

The objective, the length of catheter in cerebrospinal fluid, is the quantity the literature ties to survival in adults. In children, where age and infection dominate failure, tip position has not predicted survival[2,56,57]. In adults and mixed cohorts, tips lying free in the body or at the foramen of Monro fail less often than tips against the wall or the septum[5,8,58], with hazard ratios of 1.57 and 1.66 for suboptimal and poor placement[8] and 0.21 for a tip surrounded by cerebrospinal fluid[9]. Raising Grade 1 placement from 37% to 74% with navigation cut early revision from 22% to 5.9%[54]. The historical series here found no association[1], as expected where other causes are at work. Revision also follows infection, distal obstruction, overdrainage and valve failure[4], and in that series fixed-pressure valves were associated with more revisions[1]. Placement is the determinant that a plan can change; a Development-stage study has first to establish that the catheter can be placed where intended[53,59], and whether the placement gain translates into survival is the hypothesis for the next stage. The proximity analysis behaved as anatomy predicts. Across 47 structures the executed catheter reproduced the planned distances to within a few millimetres, the only consistent reductions being to the thalamus and caudate, which form the floor and lateral wall of the ventricle, without contact by any executed catheter. Four prior approaches frame the contribution. VPSEAR[23] mapped the posterior skull from which a perpendicular enters the ventricle but did not measure how far the ray runs inside it, and was evaluated in silico and on three-dimensionally printed skulls[55], not in patients. Automated planning of the frontal drain was evaluated in silico, on 343 CT scans[60] and, with landmarks located by a neural network, on 262 CT and MRI scans for navigated placement[61]. Automated planning for haematoma evacuation[47] aims lines at a target and was evaluated retrospectively. Duong et al.[21] chose the posterior entry on the individual CT, with 84.9% optimal placement, but with the shortest parenchymal path as the objective, an oblique direction held by eye and no depth. The present system searches for the point, takes the skull normal as the direction so that no guide is needed, and returns the depths. Every candidate is rehearsed in the twin before surgery as a catheter of its own diameter, read for $D_1$, $D_2$ and any contact with the ventricular wall, the septum or the contralateral ventricle. The executed catheter is then measured on the postoperative CT registered to the same twin, so that plan, rehearsal and measurement share one frame. A fixed landmark in practised hands can also reach 90.1% Grade 1[19]; constructed by its definition on the present cohort, that modified Keen's point enters over the temporal bone and leaves the ventricle through its medial or inferior wall in all but one case (Supplementary Table 3). Their series was 47% posthaemorrhagic and 35% normal pressure hydrocephalus, ventricles in which the atrium is wide and superficial and a catheter stopped at 5 to 6 cm sits inside it. In ventricles of the size seen here, the same offset overshoots, because a landmark is tuned to the anatomy it was developed on; only a per-patient plan can tell when the patient is not that anatomy. Compared with navigation, the Grade 1 rate here lies at the upper end of the 73 to 90% reported for navigated insertion[33,34], obtained with a ruler; navigation, ultrasound and robotic systems take a trajectory as their input, and where they exist, the computed trajectory can be that input. An augmented-reality application developed at the study institution shows the skull and ventricles on a head-mounted display registered to the head, so far on phantom skulls[62,63]; such a display would carry the computed entry and direction into the field.

This study has limitations. It is a single-centre series of 11 patients; the confidence interval on the Grade 1 rate (62 to 98%) is wide, although its lower bound is six times the historical rate. The comparison with historical controls is not randomised, and the series were graded by different readers on imaging obtained at different intervals. The Keen's point comparator follows the landmark's definition exactly, which credits the landmark with an execution freehand insertion does not achieve; the freehand series is the executed comparison. The postoperative CT was obtained within 24 hours of surgery, after drainage had begun, so the deviations in Table 1 contain early brain shift and any registration error together with the error of the hand and are an upper bound on execution error. Follow-up was limited to the immediate postoperative scan, so shunt survival, occlusion and revision were not assessed here; the cohort remains under follow-up and those outcomes will be reported when it is complete. The two lateral ventricles are separated by the mid-sagittal plane through the centroid of the inner skull, which assumes little midline shift; asymmetric or shifted anatomy requires explicit segmentation of the septum pellucidum. The scope of the study is adults with an Evans index of 0.45 or less whose ventricles satisfied the 2.5 cm floor, insertion on a hemisphere whose corridor was free of a lesion, a pinna not folded against the scalp in the scan, and thin-slice CT from one scanner. The plan does not segment the choroid plexus, and the growing skull[50,64] would need its own parameterisation. A dedicated pinna segmentation, or a positioning instruction to the radiographer at acquisition, would remove the pinna requirement.

An IDEAL Stage 2b multicentre study will enrol a larger cohort and several surgeons at each centre, so that execution fidelity can be measured per surgeon and the Grade 1 rate tested across centres and scanners, with follow-up to at least one year for the placement-to-survival hypothesis. On revision, where equipoise is real, an IDEAL Stage 3 randomised trial of computed against landmark insertion is planned, with revision at one year as the primary endpoint. Recording the catheter's position at insertion with burr-hole ultrasound or intraoperative CT, used as a study instrument, would separate the error of the hand from early shift. Learned segmentation[65–67] would extend the twin to low-contrast or distorted ventricles without changing the search. The single Grade 2 placement points to a refinement. The cannula was passed to the full planned depth, so a plan whose tip lay close to the septum was executed faithfully into contact, since nothing deflects. A minimum distance from the planned tip to the septum and ventricular walls added to the constraints would keep the rigid technique and its execution fidelity, and passing the cannula only to first cerebrospinal fluid and advancing the catheter without its stylet would let the tip follow the cavity, at the cost of some depth control. Extending the same search to a Kocher's or Frazier's window would let the twin place the three routes side by side for the individual patient. The cost of entry for a participating centre is a CT and a ruler; running the plan within the viewer the surgeon already uses would bring the emergency case, including external ventricular drainage through Keen's point[68], within reach.

## Methods

### Overview

This work has two components. The first is a planning system that constructs a static, patient-specific digital twin from the preoperative CT and returns a plan as four numbers: the two measurements A and B from the pinna, and the depths $D_1$ through parenchyma and $D_2$ within the ventricle. The second is the analysis of the preoperative and postoperative CT of 11 consecutive patients in whom that plan was executed freehand. The planning step is automatic from the DICOM series and the side of insertion to the plan. The segmentation, the alignment to the head frame and the search take no operator input, and the surgical team reviews the alignment and the segmentation in the interface before the search, where the alignment can be corrected. The geometric constructions are specified formally in Supplementary Note 1 and the search in Algorithm 1.

### Setting and eligibility

The series comprises consecutive patients who underwent posterior parietal VP shunt insertion with the plan by the neurosurgical service of Thammasat University Hospital between October 2024 and March 2026. The plan was used when a preoperative non-contrast CT of the head suitable for reconstruction was available, the pinna was not folded against the scalp, the Evans index was 0.45 or less, and the surgical team intended a posterior parietal approach, the institution's default route. Ventricular size was described by the Evans index, the ratio of the maximal frontal horn width to the maximal inner skull diameter, and by the frontal horn width itself, both measured on the preoperative CT in the axial plane through the anterior and posterior commissures.

### Digital twin construction

All preoperative CTs, and 10 of 11 postoperative CTs, were acquired on a single scanner (Spectral CT 7500, Philips, Best, the Netherlands) as helical non-contrast head studies, head-first supine, with a 512 × 512 matrix, in-plane pixel size 0.39 to 0.52 mm, 0.5 mm reconstruction interval, and standard brain kernel (UB) with iDose$^4$ level 3 iterative reconstruction. One postoperative CT (Case 7) was acquired on a second scanner (IQon Spectral CT, Philips) with the same matrix and reconstruction interval. The system takes the DICOM series of the routine preoperative CT as its only input, the thin-slice reconstruction of the routine helical acquisition, a setting available on any multidetector scanner from the same raw data at no additional radiation. Each scan is resampled to 0.5 mm isotropic voxels before segmentation, so that the morphological operations and the connected-component labelling act identically in every case and in all three directions.

The skull and the lateral ventricles are large, of stereotyped shape, and made of bone and cerebrospinal fluid whose Hounsfield ranges are distinct from everything around them, so thresholding followed by rule-based selection on size and shape separates them. The soft-tissue structures that enter the proximity analysis were labelled offline with a learned model. The skull was segmented by Hounsfield thresholding of bone, the scalp as the non-air, non-bone tissue outside it, the external ear as the scalp component lateral to the external acoustic meatus, and the ventricular system by thresholding to the CSF range within the intracranial compartment. Each mask was refined by connected-component labelling and morphological operations with a spherical structuring element (Supplementary Note 1, equation S1). The ear segmentation assumes a pinna that stands off the head, which is why a scan in which the head holder or a positioning strap folds the pinna against the scalp is not eligible.

The atrium, the junction of the body with the occipital and temporal horns, was labelled on the ipsilateral ventricle by a geometric rule in the head frame. Two coronal planes tangent to the largest circle inscribed in the most posterior axial cross-section of the ventricle divide it into the occipital horn behind, the atrium between, and the body with the frontal and temporal horns in front (Supplementary Figure 1, Supplementary Note 1).

Triangulated surfaces of the inner and outer tables, the scalp and the ventricle were extracted from the binary masks by marching cubes, and every vertex carries the unit normal of its mask, directed out of the mask. All cases were placed in a common head frame defined by MNI alignment and the Frankfurt plane, located from the porion and orbitale; the volume was resampled into that frame with third-order B-spline interpolation for the intensities and nearest-neighbour interpolation for the masks. The vertical axis of the frame is the Frankfurt normal, and the mid-sagittal plane passes through the centroid of the inner skull, perpendicular to the left-right axis; the lateral ventricle mask was divided by it into the ipsilateral and contralateral ventricles, the ipsilateral being the one on the side of the planned entry.

Because the skull mesh carries both tables, a vertex was assigned to the inner table when its normal pointed toward the centroid of the inner skull (Figure 7a). The trajectory direction at a candidate entry is that vertex's inward normal, and the same normal defines the skull normal at the executed entry used for the angular deviations in Table 1; the entry is anchored to the inner table because the two tables are not parallel and the insertion direction is established where the catheter enters the intracranial compartment.

### Trajectory generation and ranking

Candidate entries are indexed in the coordinates in which a surgeon measures Keen's point, so that the plan transfers to the patient with a ruler (Figure 7). A is the height above the top of the helix, perpendicular to the Frankfurt plane, and B is the arc length posteriorly along the scalp in the horizontal plane at that height. The scalp point so reached is carried to the inner table along the local surface normal. Each inner-table vertex is read in reverse, B running to the scalp point nearest the vertex and no further than the midline, so the window lies on the side of insertion.

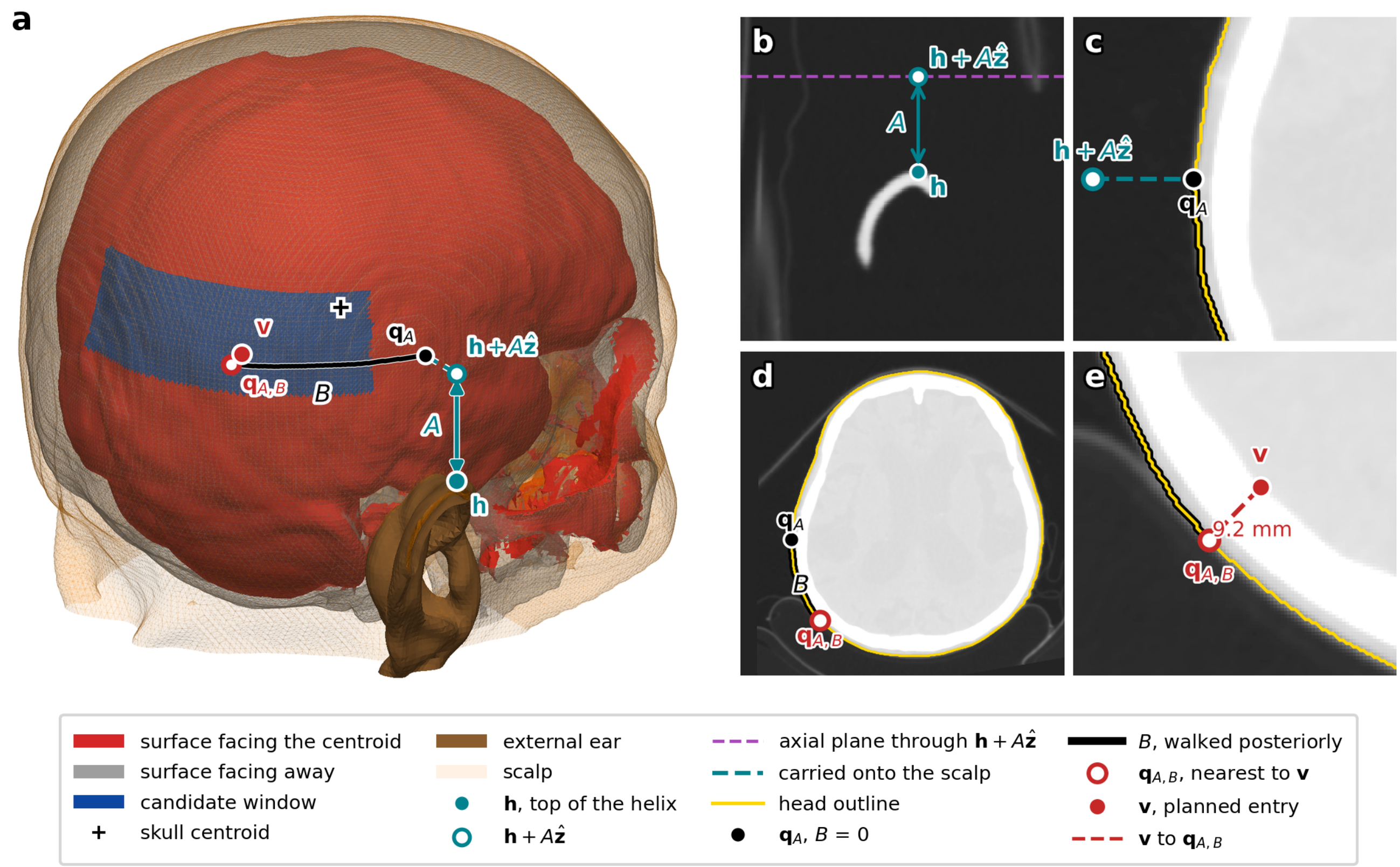

Figure 7. The inner table, the candidate window and the (A, B) coordinates of an entry, Case 8. (a) The head from behind and to the right, the scalp and the outer table translucent: the skull surface is red where the outward normal points toward the skull centroid (cross), the inner table on the vault, and grey where it points away, the outer table. The candidate window (blue) is A 2.0 to 4.5 cm above the top of the helix and B at least 2.0 cm posterior along the scalp, projected to the inner table; the skull base and the face, where the test does not separate the tables, lie outside it. The construction of panels b to e is drawn on the head: from the top of the helix h on the external ear (brown), the height A to the point $h + A\hat{z}$ in air, the carry onto the scalp at $q_A$, the walk B posteriorly along the scalp to $q_{A,B}$, and the entry v in the window. (b–e) The same construction on the preoperative CT: (b) from h, A is measured along $\hat{z}$, the upward normal of the Frankfurt plane, to $h + A\hat{z}$, in air, 2.9 cm here; (c) that point is carried to the nearest point of the head outline in the same axial plane, $q_A$, 17.4 mm away here; (d) B is the arc length along the outline from $q_A$, posteriorly, to the outline point closest to the entry, $q_{A,B}$, 5.5 cm here; (e) the entry v and its nearest outline point $q_{A,B}$, 9.2 mm apart in the plane.

Every vertex of the inner-table mesh was assigned its (A, B) by that reading, and every vertex with A between 2.0 and 4.5 cm and B of at least 2.0 cm was a candidate (Figure 7a). The window brackets the classical Keen's point, which lies at A = B = 2.5 to 3 cm in the same coordinates, and covered a median of 22.5 $cm^2$ of the inner table (range 19.1 to 24.3), 4.1% of the inner table on the side of insertion. Candidates were searched in two bands: those with B up to 4.5 cm first, and those with B beyond 4.5 cm only if no candidate in the first band satisfied every constraint. A more posterior entry becomes progressively harder to reach with the patient supine and the head turned, and far enough back would require the prone positioning of the occipital route. Supplementary Figure 2 shows one case in which no candidate in the first band satisfied every constraint, so the plan came from the second.

Algorithm 1 sets out the search. Each candidate is first tested along a ray marched along its inward normal through the label volume (Figure 8a), cheap enough that the whole window can be evaluated exhaustively, with samples in the contralateral ventricle after a septal crossing counting as overshoot. Line 7 tests five conditions. The entry must lie within the atrium label; parenchymal traversal, the track through tissue, must not exceed 4.0 cm; and intraventricular length must reach 2.5 cm, the perforated inlet segment plus a margin for the ventricle to contract once drainage is established[10,11]. Insertion depth must reach 6.0 cm, and the direction must have no posterior component in the head frame, which would carry the catheter from the atrium back into the occipital horn rather than forward along the body.

Candidates that pass are rebuilt as a three-dimensional cylinder along the ray (Figure 8b) of the catheter's outer diameter plus a safety margin. Its diameter, 3.0 mm in this study, covers the 2.5 to 2.7 mm catheters used with a radial margin of 0.25 to 0.15 mm and any catheter up to 3 mm in outer diameter, and is adjustable to the catheter and margin a centre chooses. The cylinder is advanced within the ipsilateral ventricle, bounded by its wall, the septum pellucidum and the contralateral ventricle alike, and its effective intraventricular length $D_2$ ranks the candidates and is the length reported; since a cylinder cannot be longer than its own ray, the search stops early (line 12) and remains affordable on an ordinary CPU. This check is the digital twin's rehearsal of the insertion, a straight catheter advanced through the preoperative anatomy before a plan is chosen, and the same cylinder measures every arm of Table 2 and Supplementary Table 3.

The plan is reported as (A, B) in centimetres from the pinna reference with the depths $D_1$ and $D_2$, whose sum D is the insertion depth from the inner table to the catheter tip (Table 2, Figure 1). The system runs on a CPU without a graphics processor, with optional GPU acceleration of the cylinder construction where one is present, and is compiled into a standalone application that runs on a hospital computer without a Python installation, the same build being used throughout the series (Supplementary Note 2). The interface (Supplementary Figure 3) presents the plan in the layout of a neuronavigation workstation, three orthogonal views, three trajectory-aligned views and a three-dimensional render (Figure 1b), with the insertion side, the alignment mode and the ventricle display selectable.

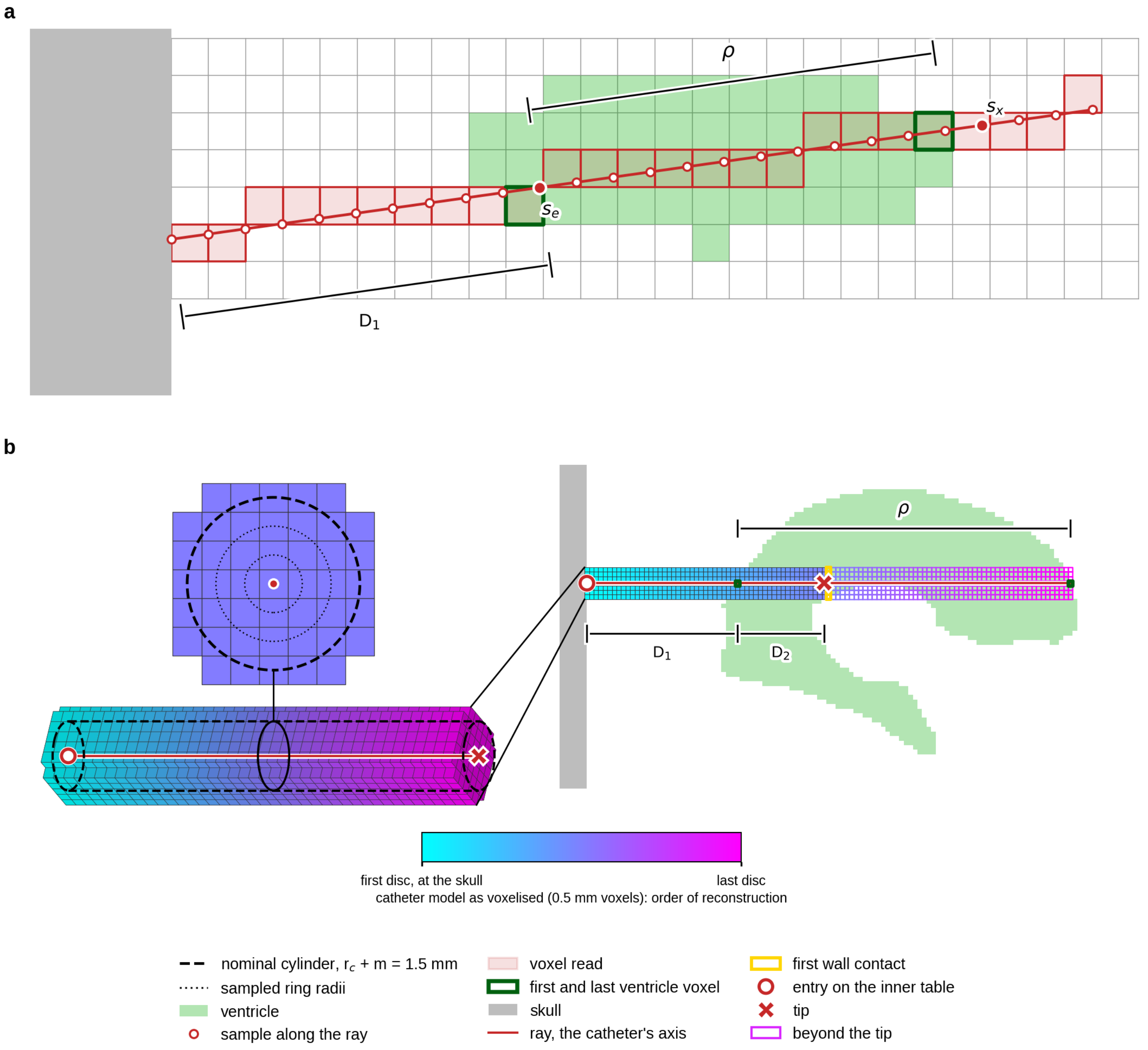

Figure 8. The ray, the catheter model and the volumetric check. (a) The one-dimensional phase: a ray marched from an inner-table vertex along its inward normal, each sample reading the label of the nearest voxel (outlined). The first and last samples inside the ventricle (green, outlined dark green) fix the intraventricular length; $D_1$ is the parenchymal traversal to the first of them, and ρ is the length between them. (b) The catheter model and the volumetric check. Left, a 20 mm rod built from discs, coloured by the order in which they are laid down, entry (open circle) to tip (cross), with one of its discs enlarged from the section marked at mid-length: the disc lies on the 0.5 mm grid, sampled on whole-voxel radii (dotted), and the nominal radius $r_c$ + m = 1.5 mm (dashed) is rounded up to whole voxels. Right, the model advanced through a longitudinal section along the axis on one patient's ventricle outline (green); the two lines join the end of the rod to its first disc at the entry on the inner table. The ray (red) fixes the entry and the exit and the rod is built only as far as the exit; the virtual catheter is advanced disc by disc and the tip is set one disc before the first disc that would meet the wall (yellow outline); discs beyond the tip are drawn as outlines. $D_2$ is the part of ρ retained after the check.

### Algorithm 1. Trajectory search

```
Input    label volume with the ipsilateral lateral ventricle Ω_V and its atrium Ω_A
         inner-table vertices v, each with its inward unit normal n and its (A, B)
         anterior axis y of the head frame
         catheter radius r_c and radial margin m
Output   (A, B), D1 and D2 of the plan (D = D1 + D2), or "no feasible trajectory"

// Phase 1: one-dimensional filter over the window
 1: C ← ∅
 2: for each vertex v with 2.0 ≤ A ≤ 4.5 cm and B ≥ 2.0 cm do
 3:     march the ray r(s) = v + s·n through the label volume
 4:     if the ray misses Ω_V then continue
 5:     s_e ← first s with r(s) ∈ Ω_V;   s_x ← first s > s_e with r(s) ∉ Ω_V
 6:     D1 ← s_e;   ρ ← s_x − s_e
 7:     if r(s_e) ∈ Ω_A and D1 ≤ 4.0 cm and ρ ≥ 2.5 cm and D1 + ρ ≥ 6.0 cm and n·y > 0
 8:         then C ← C ∪ {v}

// Phase 2: volumetric verification, first band B ≤ 4.5 cm, then B > 4.5 cm
 9: for each band in (B ≤ 4.5 cm, B > 4.5 cm) do
10:     best ← none
11:     for each v ∈ C in the band, in decreasing order of ρ do
12:         if best ≠ none and D2(best) ≥ ρ(v) then break    // no later v can exceed best
13:         build the cylinder of radius r_c + m along n from v
14:         D2 ← greatest advance beyond s_e with the cross-section wholly within Ω_V
15:         if D2 ≥ 2.5 cm and D1 + D2 ≥ 6.0 cm and (best = none or D2 > D2(best))
16:             then best ← v
17:     if best ≠ none then return (A, B) of best and D = D1(best) + D2(best)
18: return "no feasible trajectory"
```

$s_e$ and $s_x$ are the ray's entry and exit depths, $D_1$ and ρ its parenchymal traversal and intraventricular length, and $D_2$ the effective intraventricular length of the cylinder (Supplementary Note 1).

### Surgical execution

The side of insertion was chosen by the surgical team on clinical grounds before planning, as in routine practice, and entered as an input to the system, which confines its search to the inner table of that side; by clinical convention the side selected is one whose corridor and ventricle are free of a tumour, cyst or haematoma. The right side is preferred where the anatomy allows, to keep the catheter out of the dominant hemisphere. The plan sits in the routine pathway between the preoperative CT and the operation. The DICOM series is loaded on the hospital computer, the plan is computed and reviewed in the interface, and it is handed to the surgical team before surgery as A, B, $D_1$ and $D_2$, each to the millimetre. In theatre the team needs a ruler and the graduated cannula. No navigation, ultrasound, intraoperative imaging or insertion guide was used. The patient was supine with the head turned to the contralateral side. The surgical team measured A and B from the top of the helix with a ruler, marked the entry point on the scalp and made a burr hole at the mark. A Cushing brain cannula was passed perpendicular to the skull surface at the burr hole to the planned depth D to establish the track and confirm CSF, then withdrawn. The ventricular catheter was passed along the track in a single pass to depth D with its stylet in place. The depth was read on the graduations of the cannula from the level of the dura exposed in the burr hole, the inner table to which D is referenced (Figure 7e). The remainder of the procedure followed the institution's standard technique[1]. The ventricular catheters were barium-impregnated silicone catheters with a perforated inlet segment at the tip, of 2.5 mm outer diameter (Medtronic, Miethke, BMI) or 2.7 mm (Integra). The rigid cannula establishes a straight track through parenchyma, along which the catheter, stiffened by its stylet, follows, so the executed trajectory was modelled as a straight line from entry to tip, the assumption under which planned and executed trajectories are compared and which is applied identically to every landmark arm.

### Postoperative imaging and catheter reconstruction

A non-contrast CT was obtained within 24 hours of surgery, as is routine at the institution after shunt insertion. The postoperative CT was registered to the preoperative CT by an automated rigid transform computed with ANTs[69] under a mutual-information metric, and the alignment was checked on the skull in 3D Slicer with the plan hidden; the executed entry point, tip and centreline were carried into the preoperative frame, in which all quantities in Table 1 were computed. The executed catheter was segmented semi-automatically. A two-point line was placed from the inner-table entry to the catheter tip in 3D Slicer, the only manual step. The centreline was extracted as the intensity-weighted centroid of the barium-impregnated catheter, brighter than bone, in successive cross-sections perpendicular to that line, and smoothed; its arc length is the executed catheter length. The catheter mask was a tube of the catheter's diameter swept along the centreline, retained where the CT was bright. The executed tube carries no safety margin, so the margin is applied to the plan and not to what was executed.

### Outcome measures

The primary outcome was catheter placement grade on the postoperative CT according to Hayhurst et al.[54]: Grade 1, tip free-floating in CSF equidistant from ventricular walls and away from choroid plexus; Grade 2, tip touching the ventricular wall or choroid plexus; Grade 3, tip partially within parenchyma or failed ventricular cannulation. Tip location was additionally classified on the scheme of Chotsakulthong et al.[1] as good (ipsilateral frontal horn or body) or bad (septum pellucidum, third ventricle, contralateral ventricle, parenchyma).

Secondary outcomes were execution fidelity, proximity to critical structures, and a geometric counterfactual against the classical and the modified Keen's point. Execution fidelity comprised entry point deviation (distance between planned and executed entry points on the inner table), tip deviation, catheter length deviation and three angles (Supplementary Figure 5). The execution angle lies between the executed direction and the skull normal at the executed entry. The geometric angle lies between the planned direction and the skull normal at the executed entry, and equals the rotation of the normal between the planned and executed entry points. The composite angle lies between the planned and executed directions. The executed direction is the unit vector from the executed entry on the inner table to the catheter tip, the chord of the reconstructed centreline.

Proximity to critical structures was the minimum Euclidean distance from the trajectory centreline, sampled at sub-voxel intervals, to each of 47 structures (23 bilateral pairs and 3 midline structures). The structures were labelled with SynthSeg[70] on the preoperative CT for the planned trajectory and on the registered postoperative CT for the executed catheter, each label reduced to its largest connected component (for CSF and cerebral white matter, all components larger than 10 voxels). A structure was intersected, at distance zero, if any centreline sample lay within it. The temporalis muscle was labelled on the preoperative CT with TotalSegmentator[66] and tested against the extracranial course of each trajectory, from the inner table to the scalp.

Safety outcomes were intraoperative complications and, on the postoperative CT, haemorrhage along the track, contact with choroid plexus and penetration of ependyma.

### Keen's point comparators

For the counterfactual, a Keen's point trajectory was constructed on each patient's scan following the landmark's definition exactly. From the top of the helix, 3 cm was measured vertically, perpendicular to the Frankfurt plane, and projected onto the scalp, and 3 cm posteriorly along the scalp from that point, the two rulers being placed in 3D Slicer on each scan. The scalp point was projected onto the inner table along the surface normal and the trajectory taken as the inward normal. Because a surgeon inserting at Keen's point chooses the depth without knowing how much ventricle lies ahead, the Keen's point catheter was inserted to two fixed depths that represent that choice: 6 cm, the conventional depth, and 8 cm, the mean ventricular catheter length in the institutional freehand Keen's point series[1]. A sensitivity analysis repeated the comparison with the Keen's trajectory inserted to the same depth as the algorithm's plan for that patient. The modified Keen's point of Colombo et al.[19] was constructed as the authors describe it, 4 cm superior and 4 cm posterior to the external auditory meatus, by the same two-ruler construction from the centre of the ear-canal opening. It was projected onto the inner table along the surface normal with the trajectory taken as the inward normal (Figure 5), and inserted to 5 and 6 cm from the inner table, the total insertion depth they report. Parenchymal traversal, intraventricular length, overshoot beyond the ventricle and crossing of the septum pellucidum were computed for every landmark arm as for the algorithm's candidates, together with the chamber the line entered and the structure through which it left the ventricle.

### Historical controls

The comparator was the freehand Keen's point cohort of Chotsakulthong et al.[1], 213 consecutive insertions at the same institution between 2011 and 2020, graded on the Hayhurst scale and by tip location from routine postoperative CT or MRI at thick slice spacing and at variable intervals after surgery. No patient-level data from that series were used. The present cohort was graded from 0.5 mm reconstructions within 24 hours; thin-slice imaging detects contact between the catheter and the ependymal wall more readily than thick-slice imaging, so any bias from the difference in imaging runs against the present series in the Grade 1 comparison.

### Statistics and reproducibility

Continuous variables are reported as median with interquartile range and range. Depths and the two ruler measurements are given in centimetres to one decimal, as they are delivered to the surgical team; deviations, distances to structures and the dimensions of the scan and the catheter are given in millimetres. Proportions are reported with Wilson score 95% confidence intervals. The Grade 1 rate and the good-location rate were compared with the historical cohort by two-sided Fisher's exact test. Paired comparisons between the algorithm, the Keen's point trajectory and the modified Keen's point trajectory used the Wilcoxon signed-rank test. Exact and rank-based tests were chosen for the sample size; all tests were two-sided with alpha 0.05, the paired comparisons on $n = 11$ and the historical comparisons on 11 versus 213 patients. No adjustment was made for multiple comparisons. As an IDEAL Stage 2a development report, the sample size is the consecutive series of the period rather than a powered sample; the primary result is reported as a proportion with its confidence interval, and p values are given for the comparisons as descriptive statistics.

### Use of large language models

Large language models (Claude, Anthropic; Gemini, Google) were used under the authors' direction to assist with drafting and editing the text and to critique the manuscript before submission; the authors verified every statement, number and reference against the study data and take full responsibility for the content. No generative AI was used to produce data or images.

## Data availability

Per-case values for the plans and the counterfactual comparison are given in Table 2 and Supplementary Tables 2 and 3. Imaging data are not publicly available because of institutional restrictions but may be made available from the corresponding authors on reasonable request, subject to ethics approval and a data use agreement.

## Code availability

The source code of the planning system is not publicly available because it is the subject of a technology disclosure held jointly by the Agency for Science, Technology and Research (A*STAR), Nanyang Technological University and Thammasat University Hospital. The algorithm is specified in full in Methods, including Algorithm 1, and Supplementary Note 1, with every constraint and parameter stated, so that the method can be reproduced from the description. Requests for access for non-commercial verification will be considered by the corresponding authors subject to the agreement of the three institutions.

## Acknowledgements

This study received no funding.

## Author contributions

K.C. designed and built the planning system, performed the analyses and wrote the manuscript. V.Y. designed the clinical study, led the surgical team and co-wrote the manuscript. S.K. took part in the operations and collected the clinical data. X.Y. and S.Y.Y. supervised the work. All authors revised and approved the manuscript.

## Competing interests

K.C., V.Y., X.Y. and S.Y.Y. are named inventors on a technology disclosure covering the planning system, lodged with the Agency for Science, Technology and Research (A*STAR) jointly with Nanyang Technological University and Thammasat University Hospital; no patent application had been filed at the time of submission. S.K. declares no financial or non-financial competing interests, and the other authors declare no competing interests beyond the disclosure stated.

## Supplementary Note 1. Formal specification of the planning system

The constructions of Methods in symbols, in the order in which they are applied; notation follows Methods.

Segmentation. With $M_V^0$ the voxels of the intracranial compartment in the cerebrospinal fluid range, the lateral ventricle mask is

$$M_V = \left(\mathsf{LCC}(M_V^0 \ominus \mathcal{B}) \oplus \mathcal{B}\right) \bullet \mathcal{B}, \tag{S1}$$

where $\ominus$, $\oplus$ and $\bullet$ denote erosion, dilation and closing with a spherical structuring element $\mathcal{B}$, and $\mathsf{LCC}$ selects the largest connected component. The erosion detaches the lateral ventricles from the third ventricle at the interventricular foramen, the largest component is kept, and the dilation and closing restore the eroded margin and fill holes; the skull, scalp and ear masks are refined in the same way, by connected-component labelling and opening and closing with $\mathcal{B}$.

Atrium label. The atrium $\Omega_A$ is the part of $M_V$ between the two coronal planes of Methods, the largest inscribed circle being found from the distance transform of $M_V$ (Supplementary Figure 1). Reaching the temporal horn from a posterior parietal entry requires a long parenchymal path and yields a short intraventricular length, so such a trajectory exceeds the parenchymal cap or is outranked by one that runs forward along the body.

## Supplementary Figure 1. The atrium label

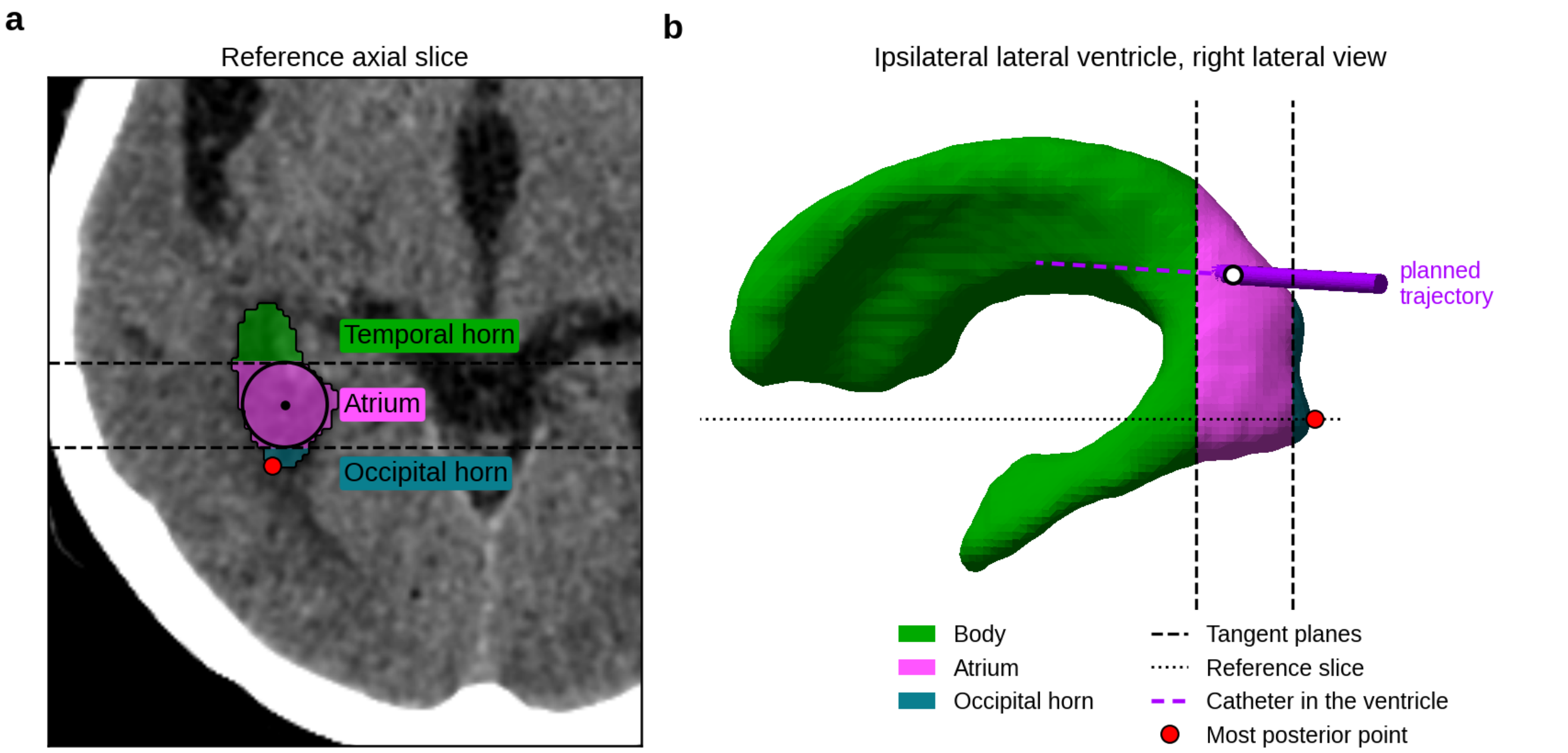

Supplementary Figure 1. The atrium label, Case 6. (a) The reference axial slice, in which the ipsilateral lateral ventricle extends furthest posteriorly (red dot). The ventricular cross-section is outlined in black with its largest inscribed circle; the dashed lines are the two coronal planes tangent to the circle. Anterior to the anterior tangent is the body with the frontal and temporal horns (green), between the tangents the atrium (magenta), and posterior to the posterior tangent the occipital horn (teal). Radiological convention. (b) The same ventricle from the right, coloured by the three labels, with the tangent planes carried through its full height (dashed), the reference slice (dotted) and the planned catheter (purple), which enters the ventricle (open circle) within the atrium and runs forward along the body; its intraventricular part is dashed. Anterior is to the left.

Ray. For a candidate vertex $\mathbf{v}$ with inward unit normal $\hat{\mathbf{n}}$ and $\Omega_V$ the ipsilateral lateral ventricle,

$$\mathbf{r}(s) = \mathbf{v} + s\,\hat{\mathbf{n}}, \qquad s \geq 0, \tag{S2}$$

$$s_e = \min\{\, s : \mathbf{r}(s) \in \Omega_V \,\}, \qquad s_x = \min\{\, s > s_e : \mathbf{r}(s) \notin \Omega_V \,\}, \tag{S3}$$

$$D_1 = s_e, \qquad \rho = s_x - s_e. \tag{S4}$$

Catheter model and effective length. The cross-section of the modelled catheter at depth $s$ is

$$\mathcal{C}(s) = \{\, \mathbf{x} : (\mathbf{x} - \mathbf{v}) \cdot \hat{\mathbf{n}} = s,\ |\mathbf{x} - \mathbf{r}(s)| \leq r_c + m \,\}. \tag{S5}$$

The cylinder is voxelised as a stack of discs whose radius is rounded up to whole voxels, so that the check errs on the side of collision (Figure 8b); $r_c + m = 1.5$ mm in this study. With $s_c = \min\{\, s \in [s_e, s_x] : \mathcal{C}(s) \subseteq \Omega_V \,\}$ the depth at which it first lies wholly within the ventricle, the effective intraventricular length is

$$D_2 = \sup\{\, t \in [0, \rho] : \mathcal{C}(s) \subseteq \Omega_V \text{ for all } s \in [s_c,\, s_e + t] \,\}, \tag{S6}$$

with $D_2 = 0$ if no cross-section fits before $s_x$, and the plan is

$$\mathbf{v}^* = \arg\max_{\mathbf{v}} D_2(\mathbf{v}), \qquad D = D_1(\mathbf{v}^*) + D_2(\mathbf{v}^*), \tag{S7}$$

Since $D_2 \leq \rho$ for every candidate, the bound that permits the early stop (Algorithm 1, line 12) holds at the voxel scale, the ray and the cylinder being read on the same axis samples (Figure 8b).

## Supplementary Figure 2. Why a plan comes from the second band

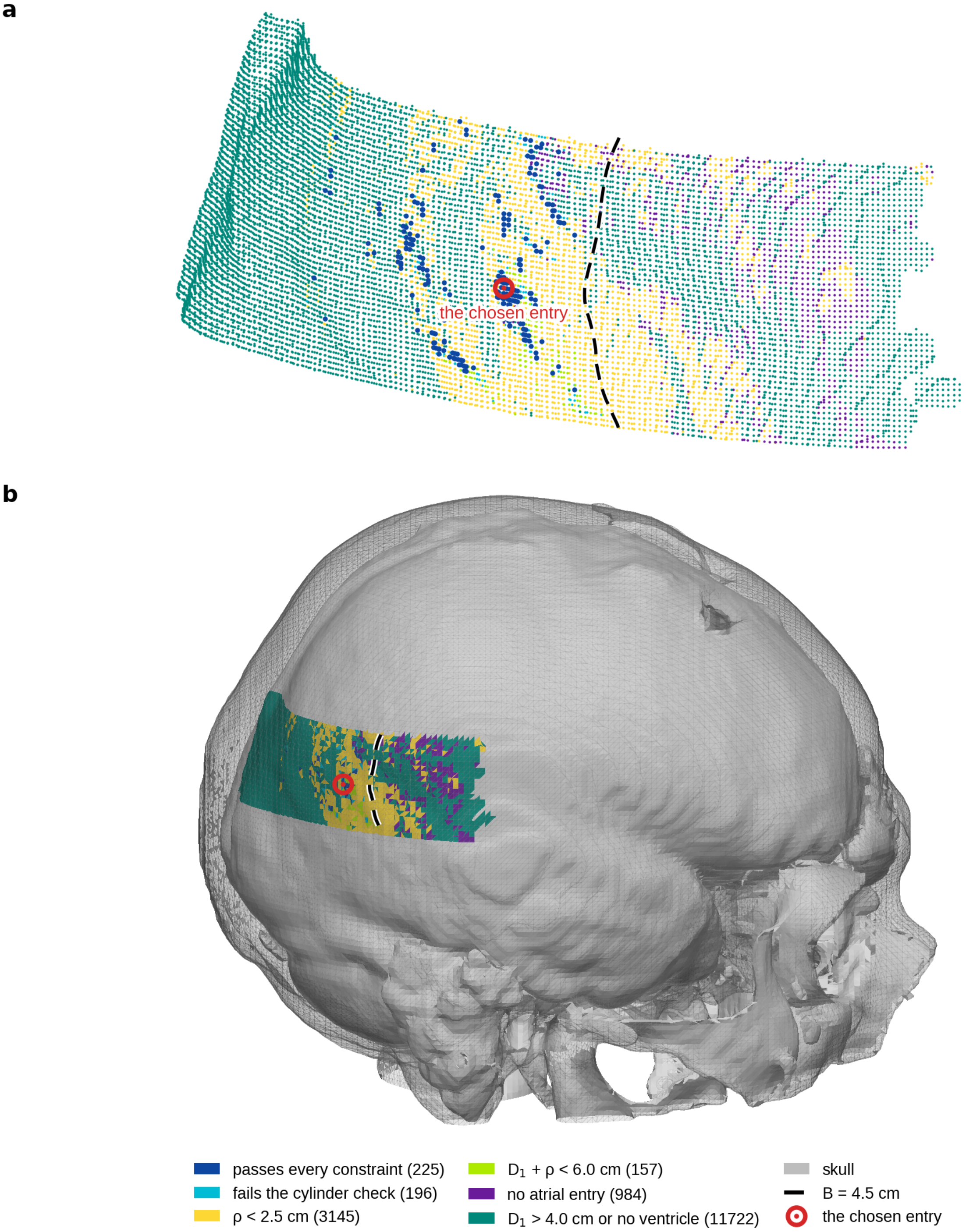

Supplementary Figure 2. Why a plan comes from the second band, Case 1. (a) The candidate window on the side of insertion, seen from the right, one mark per candidate, each coloured by the first constraint it fails, in the order $D_1 \leq 4.0$ cm, entry into the atrium, $\rho \geq 2.5$ cm and $D_1 + \rho \geq 6.0$ cm along the ray, ρ being its intraventricular length, then $D_2 \geq 2.5$ cm and $D \geq 6.0$ cm for the cylinder of the volumetric check; the larger marks satisfy every constraint. (b) The same view of the whole skull with the outer table translucent, showing where the window lies. The key serves both panels. The dashed line is B = 4.5 cm: the first band lies between it and the ear, to the right, and is searched first; the second band lies beyond it and is searched only if the first yields nothing. The number in brackets after each class in the key is its count of candidates, over the whole window of 16,429. Of the 4,654 candidates in the first band, none satisfies every constraint: 2,657 have a parenchymal traversal above 4.0 cm or a ray that never reaches the ventricle, 1,089 reach the ventricle for less than 2.5 cm, 899 enter outside the atrium, 5 have $D_1 + \rho$ below

6.0 cm, and the 4 that pass all four lose their intraventricular length to the volumetric check. The entry the plan chose (red circle) therefore lies in the second band, at B = 5.3 cm.

## Supplementary Note 2. Implementation

The system is implemented in Python 3.12.3 with NumPy 1.26.4, SciPy 1.13.0, scikit-image 0.23.2, pydicom 2.4.4 and Numba 0.59.1, with CuPy 13.1.0 for the optional GPU path, and a PyQt5 graphical interface, and is compiled with PyInstaller into a standalone Windows application. The computation times in Results were measured on a hospital computer with a laptop-class processor (Intel Core i7-12650H, six performance and four efficient cores, 16 threads), the GPU path running the cylinder construction on its NVIDIA GeForce RTX 3060 with 6 GB of memory (CUDA 12.4).

## Supplementary Figure 3. User interface of the planning system

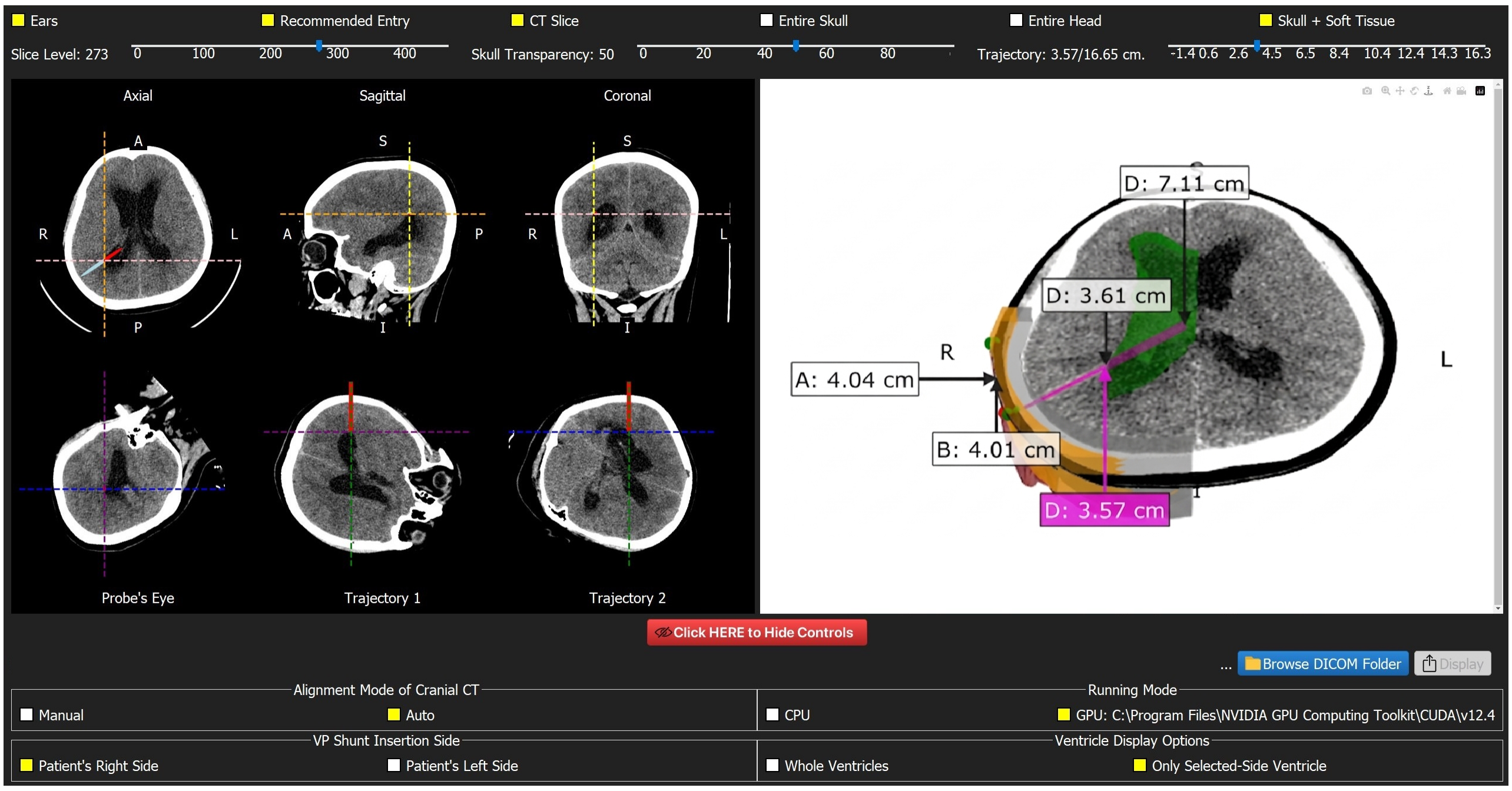

Supplementary Figure 3. User interface of the planning system, one case, identifiers removed. Left: axial, sagittal and coronal views with the planned entry, and three trajectory-aligned views, the probe's-eye view and two orthogonal views containing the trajectory. Right: three-dimensional render with the ventricle and the planned trajectory, labelled with A, B, $D_1$, $D_2$ and D. The trajectory is translucent light blue from the inner table to the ventricle, with the one-dimensional ray as a magenta line inside it, and magenta within the ventricle; the two-dimensional views show the same two segments in light blue and red, the colours of Figure 5.

## Supplementary Table 1. Direction of the planned and executed trajectories

| Case | Planned | | Executed | |
|---|---|---|---|---|
| | Azimuth | Elevation | Azimuth | Elevation |
| 1 | 47.0 | 0.7 | 51.3 | -4.3 |
| 2 | 38.9 | 4.3 | 34.4 | 6.3 |
| 3 | 52.0 | -3.5 | 57.1 | -7.3 |
| 4 | 37.5 | 5.0 | 29.8 | -2.9 |
| 5 | 47.3 | 4.5 | 50.8 | 5.2 |
| 6 | 42.7 | 2.4 | 53.9 | -2.6 |
| 7 | 48.1 | -1.4 | 54.0 | -1.6 |
| 8 | 46.5 | 1.9 | 38.2 | -0.8 |

| Case | Planned | | Executed | |
|---|---|---|---|---|
| | Azimuth | Elevation | Azimuth | Elevation |
| 9 | 38.7 | 6.2 | 35.9 | 3.2 |
| 10 | 38.6 | 10.5 | 34.0 | 12.0 |
| 11 | 49.9 | -9.9 | 50.0 | -5.9 |
| Median | 46.5 | 2.4 | 50.0 | -1.6 |
| IQR | 38.8 to 47.7 | -0.4 to 4.8 | 35.2 to 52.6 | -3.6 to 4.2 |

All values are in degrees; azimuth and elevation are drawn and defined in Supplementary Figure 4. Every planned direction satisfies the anterior requirement of Methods. All planned and executed directions point forward and inward, 30° to 57° from straight medial, and lie within 12° of the Frankfurt plane. Executed directions differed from planned by a median of +0.1° in azimuth and −2.7° in elevation, so the catheter ran a little below the plan without a systematic sideways bias.

## Supplementary Figure 4. Azimuth and elevation of a trajectory

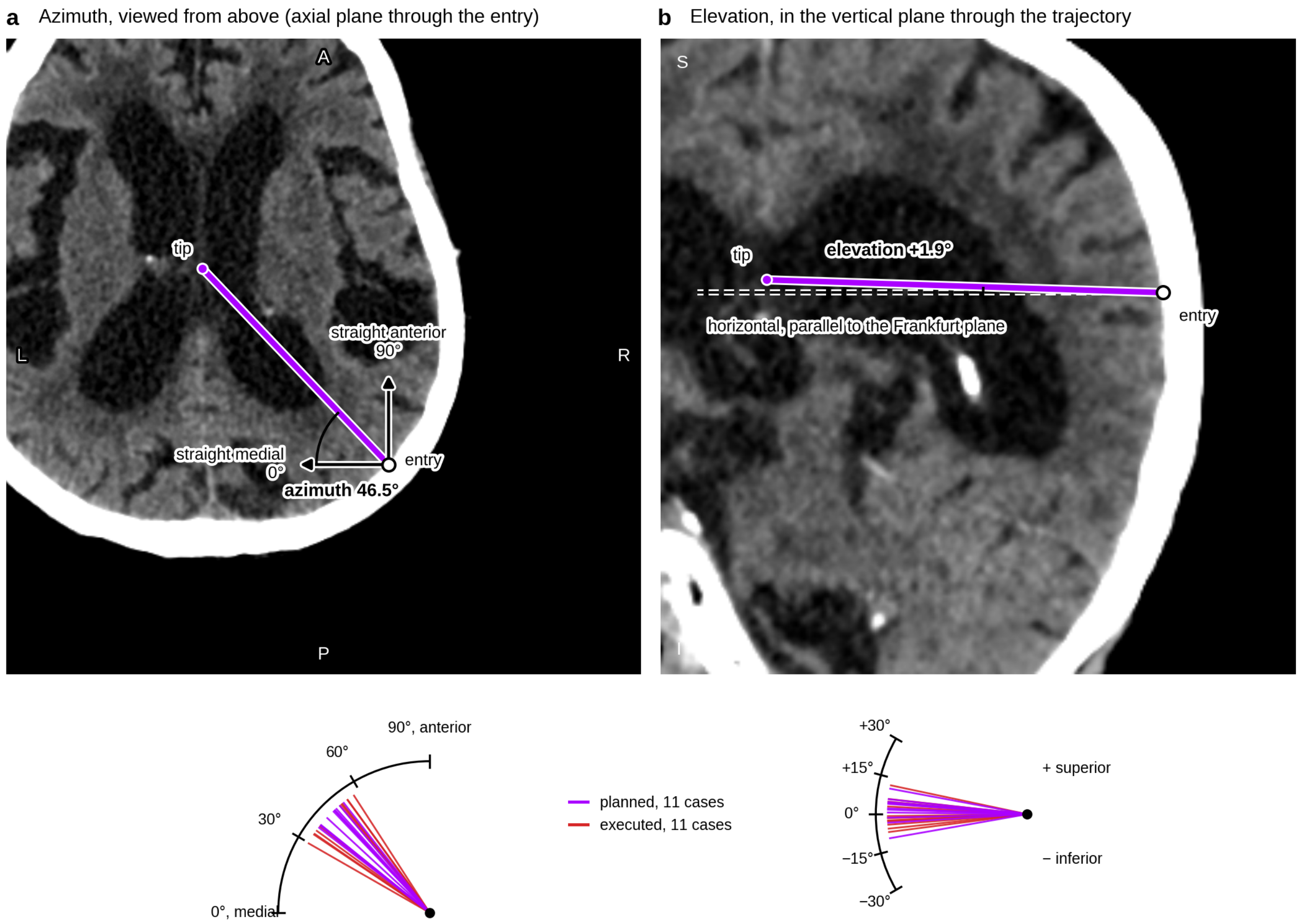

Supplementary Figure 4. Azimuth and elevation of a trajectory (Case 8, planned). (a) Axial slice through the entry, viewed from above, anterior up, right on the right. Azimuth is the angle of the trajectory from straight medial (0°) toward straight anterior (90°), measured in the plane parallel to the Frankfurt plane. (b) The CT reformatted in the vertical plane through the trajectory. Elevation is the angle of the trajectory above (positive) or below (negative) the horizontal, which is parallel to the Frankfurt plane. Below each panel, the 11 planned (purple) and 11 executed (red) values of Supplementary Table 1 as spokes on a protractor.

## Supplementary Figure 5. Deviations between planned and executed trajectories

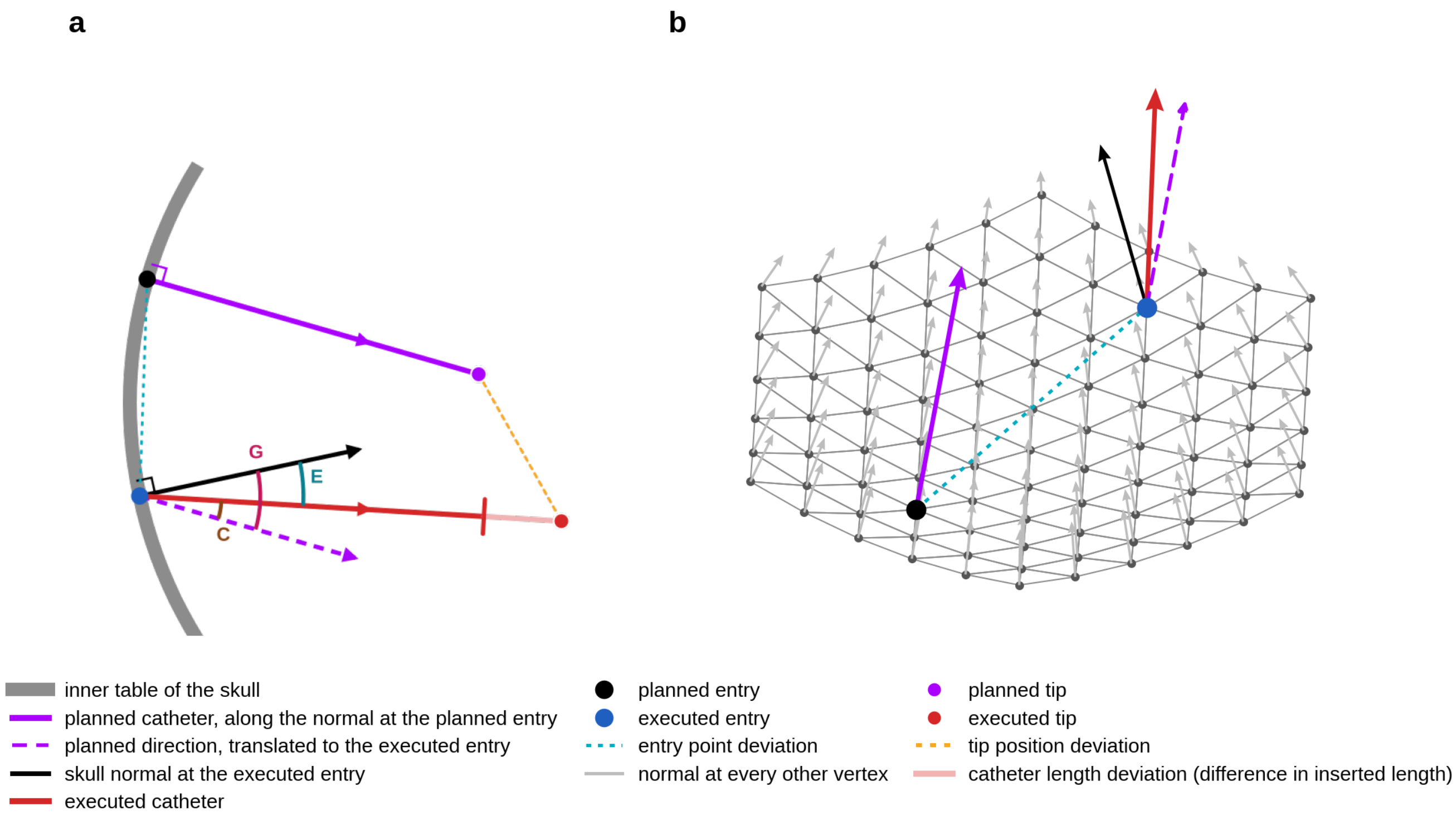

Supplementary Figure 5. Deviations between planned and executed trajectories. (a) The three angles in one plane: execution (E), the executed direction against the skull normal at the executed entry; geometric (G), the planned direction against that normal; composite (C), the planned against the executed direction. The planned direction is translated to the executed entry (dashed) so that the angles share a vertex. Entry deviation is the distance between the two entries, tip deviation the distance between the two tips, and length deviation the part of the executed catheter beyond the tick that marks the planned length. The entry shift, the angles and the length difference are exaggerated. (b) The same quantities on a patch of the inner-table mesh, where each vertex carries its own normal. The key applies to both panels. Cohort medians are given in Table 1.

## Supplementary Table 2. The plan for each case, and the catheter that was inserted, in the same coordinates

| Case | Planned | | | | | Executed | | | | |
|---|---|---|---|---|---|---|---|---|---|---|
| | A | B | $D_1$ | $D_2$ | D | A | B | $D_1$ | $D_2$ | D |
| 1 | 3.1 | 5.3 | 3.4 | 3.4 | 6.8 | 3.4 | 6.0 | 3.4 | 3.2 | 6.6 |
| 2 | 2.5 | 4.4 | 3.3 | 4.2 | 7.5 | 2.3 | 4.0 | 3.1 | 4.7 | 7.8 |
| 3 | 3.7 | 6.6 | 2.4 | 4.0 | 6.4 | 3.8 | 6.1 | 2.5 | 3.8 | 6.3 |
| 4 | 2.4 | 4.3 | 3.6 | 3.2 | 6.8 | 3.1 | 4.0 | 3.5 | 3.4 | 6.8 |
| 5 | 3.0 | 5.4 | 2.1 | 5.3 | 7.4 | 3.1 | 5.4 | 2.1 | 5.2 | 7.2 |
| 6 | 3.0 | 3.4 | 2.9 | 3.9 | 6.8 | 3.1 | 4.5 | 2.9 | 3.8 | 6.7 |
| 7 | 2.4 | 4.3 | 2.9 | 4.4 | 7.3 | 2.5 | 4.5 | 3.2 | 4.0 | 7.2 |
| 8 | 2.9 | 5.3 | 2.7 | 4.8 | 7.5 | 3.2 | 5.1 | 2.8 | 4.7 | 7.4 |
| 9 | 2.0 | 4.6 | 2.8 | 4.1 | 6.9 | 2.0 | 4.4 | 2.6 | 4.6 | 7.3 |
| 10 | 2.6 | 4.7 | 3.1 | 4.3 | 7.4 | 2.5 | 4.7 | 2.8 | 4.9 | 7.7 |
| 11 | 4.0 | 3.8 | 3.7 | 3.9 | 7.6 | 3.7 | 4.2 | 3.7 | 4.0 | 7.7 |
| Median | 2.9 | 4.6 | 2.9 | 4.1 | 7.3 | 3.1 | 4.5 | 2.9 | 4.0 | 7.2 |
| Range | 2.0 to 4.0 | 3.4 to 6.6 | 2.1 to 3.7 | 3.2 to 5.3 | 6.4 to 7.6 | 2.0 to 3.8 | 4.0 to 6.1 | 2.1 to 3.7 | 3.2 to 5.2 | 6.3 to 7.8 |

All values are in centimetres; A, B, $D_1$, $D_2$ and D are as defined in Methods. The planned $D_1$ and $D_2$ are given to the millimetre as delivered to the surgical team; Table 2 gives them as measured. The executed columns are the same five quantities for the catheter that was inserted: A and B are its entry on the postoperative CT read through the same chart; D is the distance from that entry to the tip; $D_1$ is the part of it before the ventricle and $D_2$ the part from the ventricular entry to the tip, measured against the preoperative ventricle, as the plan and every arm of Table 2 are measured. At the median the executed values differed from the planned by 0.8 mm in A, less than 0.1 mm in B, 0.0 mm in $D_1$, 0.1 mm in $D_2$ and 0.4 mm in D.

## Supplementary Table 3. Modified Keen's Point[19] Constructed by Its Definition, at 5 cm and 6 cm: Per-Case Comparison

| Case | Modified Keen's point | | | | Keen's point, 6 cm | Algorithm |
|---|---|---|---|---|---|---|
| | $D_1$ | $D_2$ † | Overshoot at 5 cm | Overshoot at 6 cm | $D_2$ | $D_2$ |
| 1 | 2.3 | 1.0 | 1.6 | 2.6 | 2.3 | 3.4 |
| 2 | 3.4 | 0.3 | 1.3 | 2.3 | 2.8 | 4.2 |
| 3 | 1.9 | 1.3 | 1.8 | 2.8 | 3.3 | 4.0 |
| 4 | 2.5 | 1.2 | 1.4 | 2.3 | 0.8 | 3.2 |
| 5 | 1.9 | 1.7 | 1.5 | 2.5 | 3.7 | 5.3 |
| 6 | 1.9 | 1.4 | 1.7 | 2.7 | 2.5 | 3.9 |
| 7 | 2.3 | 2.2 | 0.5 | 1.5 | 2.7 | 4.4 |
| 8 | 1.8 | 2.9 | 0.2 | 1.2 | 2.6 | 4.8 |
| 9 | 2.4 | 1.8 | 0.8 | 1.8 | 2.4 | 4.1 |
| 10 | 2.4 | 2.4 | 0.2 | 1.2 | 3.1 | 4.3 |
| 11 | 2.6 | 1.2 | 1.2 | 2.2 | 2.8 | 3.9 |
| Median | 2.3 | 1.4 | 1.3 | 2.3 | 2.7 | 4.1 |
| IQR | 1.9–2.5 | 1.2–2.0 | 0.7–1.6 | 1.7–2.6 | 2.5–2.9 | 3.9–4.4 |

All values are in centimetres, with the conventions of Table 2. The modified Keen's trajectory left through the medial or inferior ventricular wall into cerebral white matter in six cases, the hippocampus in three and the thalamus in one, and re-entered the ventricle after a gap in one (Case 10); none crossed the septum pellucidum. At 6 cm the tip lay in the thalamus in four cases, in the third ventricle, reached through the thalamus, in three, in the ventral diencephalon, cerebral white matter and cisternal CSF in one each, and within the lateral ventricle in one (Case 10). †Identical at 5 and 6 cm in every case; the additional centimetre adds only overshoot.